\documentclass{article}

\usepackage{PRIMEarxiv}

\errorcontextlines\maxdimen

\usepackage[utf8]{inputenc} 
\usepackage[T1]{fontenc}    
\usepackage{url}            
\usepackage{booktabs}       
\usepackage{nicefrac}       
\usepackage{microtype}      
\usepackage{fancyhdr}       
\usepackage{graphicx}
\usepackage{xcolor}

\usepackage{lineno,hyperref}
\hypersetup{colorlinks, citecolor=blue, linkcolor=red, urlcolor=green}
\usepackage{amsmath,amsfonts,amssymb}
\usepackage{graphics}
\usepackage{bm}
\usepackage{multirow}
\usepackage{algorithm}
\usepackage{algpseudocode}

\usepackage{mathtools}
\usepackage{siunitx}
\DeclarePairedDelimiterXPP\BigOSI[2]%
  {\mathcal{O}}{(}{)}{}%
  {\SI{#1}{#2}}

\title{A Bayesian Framework for learning governing Partial Differential Equation from Data}

\author{Kalpesh Sanjay More\\
  Department of Applied Mechanics\\
  Indian Institute of Technology Delhi\\
  \texttt{ama212077@am.iitd.ac.in} \\
  \And
  Tapas Tripura\\
  Department of Applied Mechanics\\
  Indian Institute of Technology Delhi\\
  \texttt{tapas.t@am.iitd.ac.in} \\
  \And
  Rajdip Nayek\\
  Department of Applied Mechanics\\
  Indian Institute of Technology Delhi\\
  \texttt{rajdipn@am.iitd.ac.in} \\
  \And
  Souvik Chakraborty \\
  Department of Applied Mechanics\\
  Yardi School of Artificial Intelligence (ScAI)\\
  Indian Institute of Technology Delhi\\
  \texttt{souvik@am.iitd.ac.in} \\
}

\begin{document}
\maketitle

\begin{abstract}
The discovery of partial differential equations (PDEs) is a challenging task that involves both theoretical and empirical methods. Machine learning approaches have been developed and used to solve this problem; however, it is important to note that existing methods often struggle to identify the underlying equation accurately in the presence of noise.
In this study, we present a new approach to discovering PDEs by combining variational Bayes and sparse linear regression. The problem of PDE discovery has been posed as a problem to learn relevant basis from a predefined dictionary of basis functions. To accelerate the overall process, a variational Bayes-based approach for discovering partial differential equations is proposed. To ensure sparsity, we employ a spike and slab prior. We illustrate the efficacy of our strategy in several examples, including Burgers, Korteweg-de Vries, Kuramoto Sivashinsky, wave equation, and heat equation (1D as well as 2D). Our method offers a promising avenue for discovering PDEs from data and has potential applications in fields such as physics, engineering, and biology.
\end{abstract}


\section{Introduction}
Partial differential equations (PDEs) \cite{strauss2007partial, wazwaz2002partial} are crucial in explaining a wide range of physical phenomena. These equations are derived from the fundamental laws of physics and play an essential role in modelling and predicting various physical systems. They find their application in numerous domains, such as fluid dynamics \cite{helal2002soliton}, solid mechanics \cite{roubivcek2013nonlinear}, electromagnetics \cite{rabczuk2019nonlocal}, and quantum mechanics \cite{purohit2010fractional, howe1980quantum}. However, in reality the true governing equations are often unknown, making it challenging to predict the system's behaviour accurately. Identifying the governing PDEs from observed data has been a long-standing problem in scientific disciplines like physics \cite{courant2008methods}, engineering \cite{debnath2005nonlinear}, and biology \cite{leung2013systems}.
Traditionally, the governing PDE of any the system is constructed from first principles which is used to explain the system's behavior. However, this approach is time-consuming and requires good domain expertise. For complex systems, the principles behind the system are not always fully understood, making it challenging to construct accurate models.
Recently, data-driven approaches have gained considerable attention in identifying PDEs. These methods aim to discover the underlying dynamics of a system directly from data without prior knowledge of the governing equations. The idea behind these methods is to use observed data to construct a model that accurately describes the system's behaviour.

One of the most popular data-driven approaches for identifying PDEs is sparse identification of nonlinear Dynamics (SINDy) \cite{brunton2016discovering}. The concept is to create a list of candidate functions that could define the PDE and then use sparse linear regression to choose only the significant ones to develop the most appropriate model.
SINDy finds applications in many areas, including sparse identification in biology \cite{mangan2016inferring}, chemistry \cite{hoffmann2019reactive, bhadriraju2019machine}, fluid mechanics \cite{loiseau2018constrained, loiseau2018sparse}, identifying structures that exhibit hysteresis \cite{lai2019sparse, li2019discovering}, predictive control sparse identification \cite{kaiser2018sparse}, identifying structural dynamical systems from limited data \cite{schaeffer2020extracting},
identifying a model by recovering the underlying differential equations using time-series data of the impulse response, even when the available data is limited in duration \cite{stender2019recovery}.
However, a standard limitation of the SINDy algorithm is its sensitivity to noise. As the noise level in the data increases, SINDy may provide inaccurate results, which can be a significant problem in many practical applications.

Feature-weighted, iteratively refined, and doubly regularised sparse identification of nonlinear dynamics (FIND-SINDy) \cite{rudy2017data} is an extension of SINDy \cite{brunton2016discovering} that improves its performance on noisy or incomplete data. FIND-SINDy uses a feature weighting scheme to give more importance to certain variables in the PDE and employs a regularisation technique to stabilise the solution. FIND-SINDy has been successfully applied to discover PDEs governing various physical systems, including fluid dynamics \cite{hao2021data}, chemical kinetics \cite{ai2022study}, and for discovering the underlying equations governing
Perovskite Solar-Cell degradation \cite{naikdiscovering}. It was also used to discover the PDEs that govern a biological system from noisy experimental data. 
Despite its potential to identify governing equations from data, the algorithm's computational complexity can become a bottleneck when working with a significant amount of data. 
PDE may also be identified via the application of convolutional neural networks and symbolic networks \cite{long2019pde}, both of which can generate long-term predictions if the structure of the PDE is known.
Another relevant work in this area is \cite{fuentes2021equation}, which employs a Relevance Vector Machine (RVM) framework to discover nonlinear dynamical systems. The RVM approach is based on probabilistic modelling; it relies on sparsity-promoting priors to learn compact representations of the data, making it particularly suitable for applications where computational efficiency is critical.

The Bayesian approach is a promising data-driven technique for discovering nonlinear stochastic dynamical systems with Gaussian white noise, as described in \cite{tripura2023sparse}. This method employs an MCMC-based Gibbs sampling algorithm, which is iteratively run to identify the dynamics. Due to the robustness of system identification, it has found applications in model agnostic control \cite{tripura2022model} and digital-twin \cite{tripura2023probabilistic}. While the MCMC-based method is widely regarded as the gold standard for such applications, its computational cost can become prohibitive for larger datasets.
However, \cite{mathpati2023mantra} presents a highly efficient variational Bayes approach to address this limitation, applied to discovering stochastic differential equations from data. By leveraging the principles of variational inference, this method offers a computationally tractable alternative that can deliver comparable results to the MCMC-based approach. However, to date, the development of the Bayesian equation discovery framework is limited to discrete dynamical systems governed by ordinary differential equations; this is because of the huge computational cost associated with the Bayesian learning approach.

In this paper, we suggest a new approach for discovering PDEs from noisy data. The proposed approach exploits variational Bayes and spike-and-slab prior. While spike-and-slab prior ensures a sparse solution, variational Bayes addresses the bottleneck associated with computational cost. The proposed approach is an improvement over previous work like SINDy and FIND-SINDy. To apply the variational Bayes algorithm with a spike-and-slab prior for discovering PDEs, we first initialise the algorithm using ridge regression \cite{rudy2017data} and then use the VB algorithm to discover the model. We test the efficacy of the proposed algorithm by applying it to discover PDEs in several physical systems, including the Burgers, Korteweg-de Vries, Kuramoto Sivashinsky, wave equation, and 1D-2D heat equation. The proposed method has the following advantages:
\begin{itemize}
    \item \textbf{Computational Efficiency:}
    Variational Bayes is a computationally efficient approach that can handle large datasets with high-dimensional inputs. This is because it employs a posterior distribution approximation, which is simpler to calculate than the actual posterior distribution. When compared to previous approaches \cite{brunton2016discovering, rudy2017data, liu2021automated}, the VB framework is more scalable to equation identification.
    \item \textbf{Handles noise more effectively:}
    The variational Bayes method is intended to deal with noisy data. It employs a probabilistic approach that accounts for uncertainty in data and model parameters. This paradigm makes it easier to handle noisy data and increases the likelihood of discovering correct equations that generalize effectively with newly collected data.
\end{itemize}

The rest of the paper is organized as follows. Section \ref{prob_state}, discusses the problem statement; the proposed approach is presented in Section \ref{prob_formulation}; Section \ref{problems} entails numerical studies that demonstrate the robustness and efficacy of the proposed approach,
and Section \ref{conclusion} contains the conclusion of the study.

\section{Problem statement}\label{prob_state}
We consider a bounded domain $\Omega$ with the boundary $\partial \Omega$. With an initial condition $u_0(x,y)$ and a boundary condition $u_{\partial \Omega}(t)$ the generalized form of a partial differential equation (PDE) can be expressed as,
\begin{equation}
    \label{Gen_PDE}
    \begin{aligned}
        & F \left(x, y, t, u, u_x, u_y, Y, u_{xx}, u_{yy},u_{tt}, u_{xxx}, u_{yyy},u_{ttt}, \ldots \right) = 0, \\
        & u(0,x,y) = u_0(x,y), \\
        & u(t,\partial \Omega) = u_{\partial \Omega}(t),
    \end{aligned}
\end{equation}
where $x \in \mathbb{R}$, $y \in \mathbb{R}$, and $t \in \mathbb{R}$ represent spatial and temporal variables, respectively, $u \triangleq u(t,x,y) \in \mathbb{R}$ is the unknown function that describes the spatio-temporal evolution of the system, and the remaining expressions denote the different partial derivatives of the function $u$ with respect to its variables $x$, $y$, and $t$.

The objective of learning governing PDEs from data is to uncover the interpretable functional form of $F(\ldots)$ and determine the coefficients or parameters that govern the behaviour of the system being studied. This entails utilizing available data, such as time series measurements of the system. Time series measurements can be obtained using two different approaches: Eulerian, where sensors remain stationary in fixed locations, or Lagrangian, where sensors move along with the dynamics being measured. These measurements are crucial in inferring the underlying physical laws that dictate the behaviour of the system. However, this task of discovering PDE can be challenging in practice due to the inherent complexity of PDEs and the high-dimensional nature of the data.

\section{Methodology for Bayesian learning of partial differential equation from data}\label{prob_formulation}
In this section, we will introduce the framework of our proposed approach for solving the problem outlined in Section \ref{problems}. Our approach leverages the power of Bayesian statistics in combination with a sparse learning algorithm to discover the partial differential equation governing the system under study from the time series data collected on the spatial domain.

\subsection{Partial differential equation discovery}
Without loss of generality, we rewrite the PDE in Eq. \eqref{Gen_PDE} as,
\begin{equation}
    \bm u_t = {\bf{N}}(\bm u,\bm u_x,\bm u_{xx},\bm u_y,\bm u_{yy}\ldots, x,y),\label{pde1}
\end{equation} 
where $\bf{N}()$ in general is a non-linear function of $u(t,x,y)$ and its derivatives. Subscripts denote partial differentiation in time or space. With this setup, our aim is to discover terms of $\bf{N}()$ given that we have the time series measurements of the system at a fixed number of locations in space described by ${x}$ and ${y}$. A reasonable assumption is that the $\bf{N}()$ contains only a few terms. Thus, given the large collection of candidate basis functions, we use sparse linear regression to determine which functions are contributing to the dynamics of the system.

We measure the solution $u(t,x,y)$ over a discretized domain $D \in \mathbb{R}^{{N_x}\times{N_y}}$, where $N_x$ and $N_y$ are the numbers of points in space. We assume that we have $N_t$ numbers of temporal snapshots of the solution $u(t,x,y)$ over a discretized domain $D$. Next we perform a vectorization transformation over the complete solution as $\mathbf{u}(t,x,y) \in \mathbb{R}^{N_x \times N_y \times N_t} \to \boldsymbol{u}(t,x,y) \in \mathbb{R}^{N_x N_y N_t}$. Let $\bm{u} \in \mathbb{R}^N$ be the resulting vector where $N \leq N_x N_y N_t$. To be able to learn an expression of $\mathbf{N}$, we express the right-hand side of the Eq. \eqref{pde1} as a weighted linear combination of certain partial derivatives, which we refer as the candidate basis functions.
Once the observations for $u(t,x,y)$ are obtained, these partial derivatives are computed using numerical derivatives. We stack the candidate functions in a dictionary $\mathbf{D} \in \mathbb{R}^{N \times K}$, where $K$ is the total number of basis functions. The PDE evolution in terms of dictionary terms can be expressed as,
\begin{equation}\label{pde2}
    \boldsymbol{Y} = \mathbf{D}{\Phi} + \boldsymbol{\varepsilon},
\end{equation}
where $\boldsymbol{Y} \in \mathbb{R}^{N}$ represent the $N$-dimensional target vector, ${\Phi = [\phi_1, \ldots, \phi_K]}$ is the parameter vector, and $\bm{\varepsilon} \in \mathbb{R}^{N}$ represent the residual error vector that signifies the model mismatch error. The objective in Bayesian inference is to determine the posterior distribution of the parameter vector ${\Phi}$ based on the measurements $\boldsymbol{Y}$. To estimate the posterior distribution, one can use the Bayes formula as shown in Eq. \eqref{bayes_ut},
\begin{equation}
    p({\Phi} \mid \boldsymbol{Y})=\frac{p({\Phi}) p(\boldsymbol{Y} \mid {\Phi})}{p(\boldsymbol{Y})},\label{bayes_ut}
\end{equation}
where $p({\Phi} \mid \boldsymbol{Y})$ represents the posterior distribution, $p({\Phi})$ represents the prior distribution, $p(\boldsymbol{Y} \mid {\Phi})$ represents the likelihood function, and $p(\boldsymbol{Y})$ represents the marginal likelihood or evidence. The measurement error $\bm{\varepsilon}$ is modelled as an i.i.d Gaussian random variable with zero mean and variance $\sigma^{2}$. The likelihood function can be expressed as follows:
\begin{equation}
    \boldsymbol{Y} \mid {\Phi}, \sigma^{2} \sim \mathcal{N}\left(\mathbf{D} {\Phi}, \sigma^{2} \mathbf{I}_{N \times N}\right),
\end{equation}
where $\mathbf{I}$ stands for the identity matrix of size $N \times N$. To effectively discover the governing physics of a system, it is important that the resulting model is easily interpretable. This means that the solution of weight vector ${\Phi}$ should contain only a limited number of terms in the final model. This can be achieved by utilizing various techniques and methodologies to promote sparsity in the weight vector, ultimately resulting in a simpler and more understandable model of the system. Here, we employ the spike-and-slab (SS) prior distribution to encourage sparsity in the solution. Combining a Dirac-delta spike at zero and a continuous distribution (slab over real line), the SS prior is a hierarchical mixture prior. The SS prior can be expressed mathematically as follows,
\begin{equation}\label{spike_slab}
    p({\Phi} \mid \boldsymbol{Z})=p_\text{slab}\left({\Phi}_{k}\right) \prod_{i, \boldsymbol{Z}_{i}=0} p_{\text{spike}}\left({\Phi}_{i}\right).
\end{equation}
When applying the SS prior, the components of the weight vector that fall into the spike category are assigned a value of zero, while the components that fall into the slab category are allowed to take non-zero values. This leads to a sparse weight vector where many components are forced to be zero, which can help to prevent overfitting and improve the generalization of the model. The classification of each component is controlled by an indicator variable $\boldsymbol{Z}_i$, where if $\boldsymbol{Z}_i$ equals 1, the weight falls into the slab category; otherwise, it takes a value of 0, indicating that it belongs to the spike category.
The group of weight vectors that contains only those variables from ${\Phi}$ for which $\boldsymbol{Z}_i$ equals 1 is denoted by ${\Phi}_k$. Essentially ${\Phi}_k$ is a subset of ${\Phi}$, that includes only the weight components that are classified as a slab.
The distributions of spikes and slabs are defined as, $p_{\text {spike }}\left({\Phi_{i}}\right)=\delta_{0}$ and $p_{\text {slab}}\left({\Phi}_{k}\right)=\mathcal{N}\left(\mathbf{0}, \sigma^{2} v_{s} \mathbf{I}_{k \times k}\right)$. The random variables $\sigma^{2}$ and ${Z}$ are represented as:
\begin{equation}\label{bern}
    \begin{aligned}
    p(\boldsymbol{Z}_{i} \mid p_{0}) &=\operatorname{Bern}
    (p_{0}) ; \; i=1 \ldots K, \\
    p(\sigma^{2}) &=\mathcal{I} \mathcal{G}(\alpha_{\sigma}, \beta_{\sigma}),
    \end{aligned}
\end{equation}
where the constants $p_0 \in \mathbb{R}$, $v_{s} \in \mathbb{R}$, $a_{\sigma} \in \mathbb{R}$, and $b_{\sigma} \in \mathbb{R}$ serve as the hyperparameters in the model. The random variables ${\Phi}, \boldsymbol{Z}$ and $\sigma^{2}$ joint distribution is given as,
\begin{equation} 
    \begin{aligned}
        p({\Phi}, \boldsymbol{Z}, \sigma^{2} \mid \boldsymbol{Y}) &
         \propto p(\boldsymbol{Y} \mid {\Phi}, \sigma^{2}) p({\Phi} \mid \boldsymbol{Z}, \sigma^{2}) p(\boldsymbol{Z}) p(\sigma^{2}).
    \end{aligned}\label{eq:Bayes}
\end{equation}
Bayes formula can be used to express the joint distribution of random variables as $p\left({\Phi}, \boldsymbol{Z}, \sigma^{2} \mid \boldsymbol{Y}\right)$. The likelihood function, $p\left(\boldsymbol{Y} \mid {\Phi}, \sigma^{2}\right)$, represents the probability of observing the data, given the weight vector ${\Phi}$ and the noise variance $\sigma^{2}$. 
The prior distribution for ${\Phi}$, given $\boldsymbol{Z}$ and $\sigma^{2}$, is denoted by $p\left({\Phi} \mid \boldsymbol{Z}, \sigma^{2}\right)$.
The prior distribution for the latent vector $\boldsymbol{Z}$ is represented by $p(\boldsymbol{Z})$, while $p\left(\sigma^{2}\right)$ denotes the prior distribution for the noise variance, and the marginal likelihood or evidence is expressed as $p(\boldsymbol{Y})$.
We employ the variational Bayes (VB) approach to approximate the joint posterior distribution $p\left({\Phi}, \boldsymbol{Z}, \sigma^{2} \mid \boldsymbol{Y}\right)$. Readers are directed to  \ref{appendixA} for derivations pertaining to VB.


\begin{figure}[ht!]
    \centering
    \includegraphics[width=0.9\textwidth]{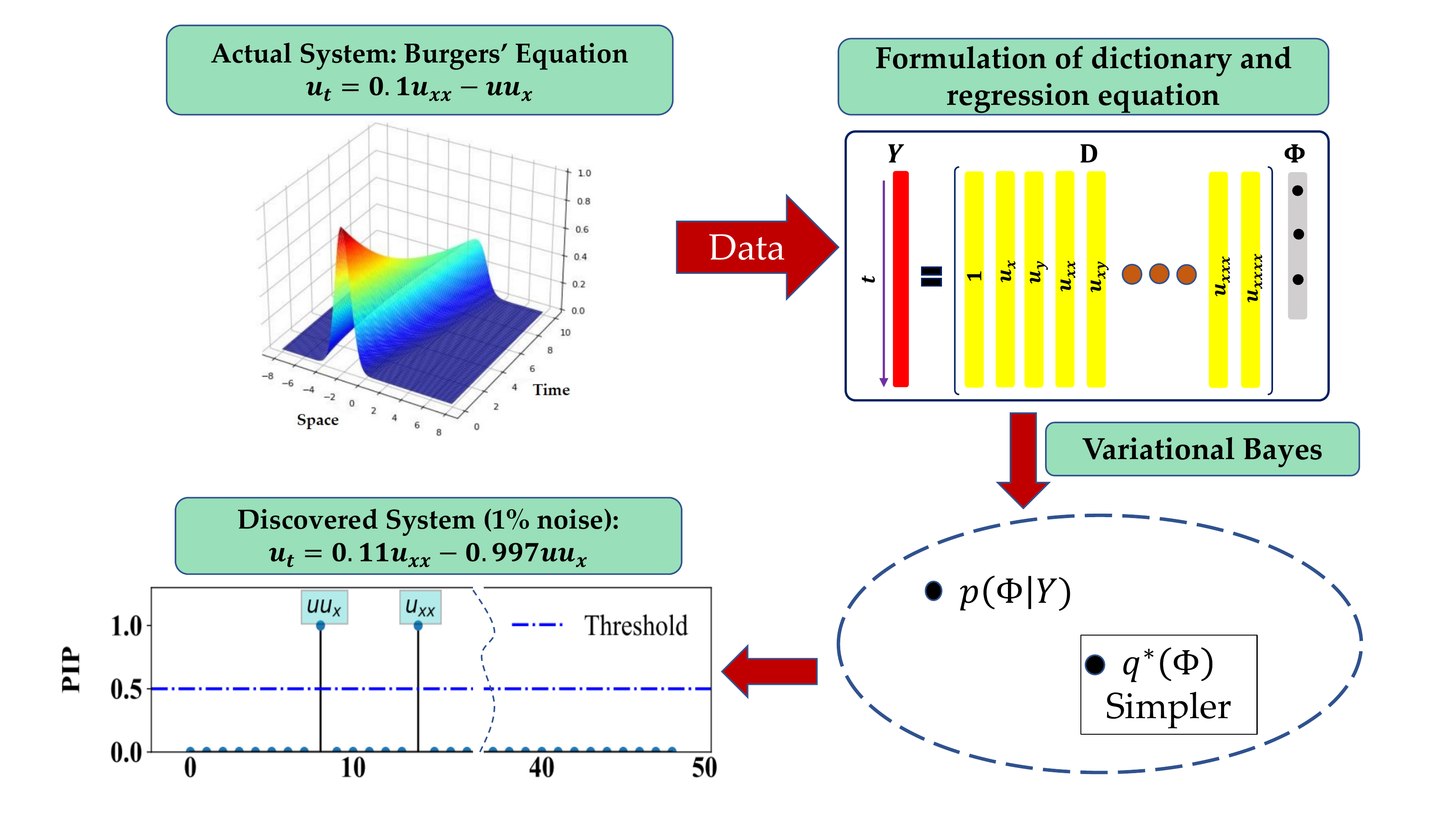}
    \caption{\textbf{Flowchart of the proposed approach}}
    \label{fig:FlowChart}
\end{figure}

\subsection{Dictionary and derivative}
The dictionary $\mathbf{D} \in \mathbb{R}^{N \times K}$ in Eq. \eqref{pde2} is a function of $\bm{u}$, derivatives of $\bm{u}$ with respect to space, and their combinations, where $K$ represents the number of basis functions in the dictionary, and expressed as,
\begin{equation}\label{dict}
	{\bf{D}} = \left[ \begin{array}{*{20}{c}}{\bf{1}}&\bm{u}_x&\bm{u}_{y}&\bm{u}_{xx}&\bm{u}_{yy}&\bm{u}\otimes \bm{u}_x & \ldots 
	\end{array} \right].
\end{equation}
Subscripts in Eq. \eqref{dict} indicate differentiation, $\bf{1}$ is a vector of ones, and $\otimes$ denotes Hadamard (element-wise) product. 
The final selection of a model is based on the marginal posterior inclusion probability (PIP) denoted as $p\left(\boldsymbol{Z}_{i}=1 \mid \boldsymbol{Y}\right)$. This probability shows how likely it is that a certain basis function will be part of the final model.
The selection process involves choosing the basis functions with a PIP value greater than 0.5.
The estimated mean and covariance of the weight vectors $\boldsymbol{\Phi}$, denoted by $\hat{\boldsymbol{\mu}}_{\Phi}$ and $\hat{\boldsymbol{\Sigma}}_{\Phi}$, are calculated by populating them with the values of $\boldsymbol{\mu}^{*}$ and $\boldsymbol{\Sigma}^{*}$ at the respective indices that correspond to the selected basis variables. 
The remaining entries of $\hat{\boldsymbol{\mu}}_{\Phi}$ and $\hat{\boldsymbol{\Sigma}}_{\Phi}$ are set to zero, as shown below:
\begin{align}
    \hat{\boldsymbol{\mu}}_{{\Phi}} &= \begin{cases}
    \boldsymbol{\mu}^{*} & \text{at selected indices,} \\
    \mathbf{0} & \text{otherwise.}
    \end{cases} \
    & \hat{\boldsymbol{\Sigma}}_{{\Phi}} &= \begin{cases}
    \boldsymbol{\Sigma}^{*} & \text{at selected indices,} \\
    \mathbf{0} & \text{otherwise.}
    \end{cases}
\end{align}

One key component of the proposed approach is to compute the derivatives present in the dictionary. We experimented with several numerical differentiation techniques and observed that the second-order finite difference \cite{leveque2007finite, thomas2013numerical} works well for clean data. However, the choice of grid spacing is critical when dealing with noisy data using finite difference techniques. The resultant $d^{\text{th}}$ derivative will have the noise of approximately $\BigOSI{}{\epsilon h^{-d}}$ if the grid spacing is $\BigOSI{}{h}$ means the $h$ number of step size and the noise has an amplitude of $\BigOSI{}{\epsilon}$. As a result, the impact of noise can greatly influence the numerical derivatives, resulting in erroneous findings and potentially severe inaccuracies.
Polynomial interpolation \cite{barthelmann2000high, bruno2012numerical} is an incredibly powerful technique that is often employed to estimate a function by using a polynomial equation. This approach proves to be particularly useful in scenarios where data is noisy and there is a pressing need to extract precise and accurate information from it.
To estimate the derivatives of each data point, a polynomial equation of degree $p$ is fitted to more than $p$ data points. Once the polynomial equation is determined, its derivatives are then utilized to estimate the derivatives of the original numerical data. In this work also, polynomial interpolation was used for computing the derivatives.

For ease of readers, the overall algorithm proposed is shown in Algorithm \ref{algvecvec}. Furthermore, a schematic representation of the proposed approach is elucidated in Fig. \ref{fig:FlowChart}.
\begin{algorithm}[ht!]
	\caption{Variational Bayesian Inference Algorithm for Learning PDE  from Data}\label{algvecvec}
	\begin{algorithmic}[1]
	\Require{The time series data collected over a spatial domain}: $\mathbf{u} \in \mathbb{R}^{N_x \times N_y \times N_t}$
	    \State{Create a target vector $\bm{Y}$ and a dictionary ${\mathbf{D}}$ using the candidate basis functions.}
	    \State{Assign an initial value to $\tau$ and initialize the variational parameter $\boldsymbol{w}^{q}$ using the ridge algorithm.}
		\While {$\operatorname{ELBO}^{(j)}-\operatorname{ELBO}^{(j-1)} < \rho$}
            \State{Update latent variable vector $Z_{i}$.}\label{step1} \Comment{Eq.\ \eqref{eq:variationa_param}}
            \State{Update slab variance $v_{s}$.}
            \State{Update the noise prior $a_{\sigma}^{q}$, $b_{\sigma}^{q}$.}
            \State{Update the initialize vector $\tau$ and variational parameter $\boldsymbol{w}^{q}$.}
            \State{Update weight vector $\Phi$.}
		\State {Calculate the ELBO using the variational parameters for the $j$th iteration.}\label{stepn} \Comment{Eq.\ \eqref{eq:elbo}}
        \State{Repeat steps \ref{step1}$ \to $\ref{stepn}.}
		\EndWhile
	\State{Compute marginal PIP values for each basis function, given the observed data $\bm{Y}$.} \Comment{Eq.\ \eqref{bern}}
	\State{Choose the basis functions for the final model that have a PIP value greater than 0.5.}
	\Ensure{The mean ${\bm{\hat{\mu}}}_{\Phi}$, covariance ${\bf{\hat{\Sigma}}}_{\Phi}$ and PIPs $p(Z_k=1|\bm{Y})$ for $k=1,\ldots,K$.}
	\end{algorithmic}
\end{algorithm}

\section{Numerical problems}\label{problems}
In this section, the proposed method is applied to five different numerical problems that are commonly encountered in practice, demonstrating the approach's robustness and effectiveness. The numerical problems studied include (a) the Heat equation (1D and 2D), (b) the 1-D Burgers equation, (c) the Korteweg-de Vries equation, (d) the Kuramoto Sivashinsky equation, and (e) the 1-D Wave equation.

\subsection{Heat equation (1D)}\label{sec:Heat}
The first example is the 1-D Heat equation. In the 1D heat equation, the temperature of an object depends on only one dimension, and heat doesn't flow out of the object's side.
The heat equation is used in a wide range of fields, including heat transfer, fluid dynamics, and even geotechnical engineering. It is also used in probability and to describe random walks, which is why financial math uses it. The equation of 1D heat is as follows,
\begin{equation}\label{1d_heat_eq_1}
    \begin{aligned}
        &u_t = \alpha  u_{xx}, \; x\in[0,1], \; t\in[0,0.2],\\ 
        &\text{IC :} \; u(x,0) = \sin{(\pi x)},\\
        &\text{BC: } u(0,t) = 0, \; u(1,t) = 0,
    \end{aligned}
\end{equation}
where $u(x,t)$ is the unknown function to be solved, $t$ is time, and $x$ is the coordinate in a space. The diffusion coefficient $\alpha = {K_o}/{c\rho}$ determines how fast $u$ changes in a time, where $K_o \in \mathbb{R}$ is the thermal conductivity, $c \in \mathbb{R}$ is specific heat, and $\rho \in \mathbb{R}$ is the density.
For illustrating the application of the proposed approach, we generated synthetic data by solving Eq. \eqref{1d_heat_eq_1} using finite difference method. The spatial domain is discretized into 44 points and a time-step of $\Delta t = 6.6 \times 10^{-6}$s is considered with this setup, the objective is to identify the governing PDE from data.
\begin{figure}[!b]
    \centering
    \includegraphics[width=0.8\textwidth]{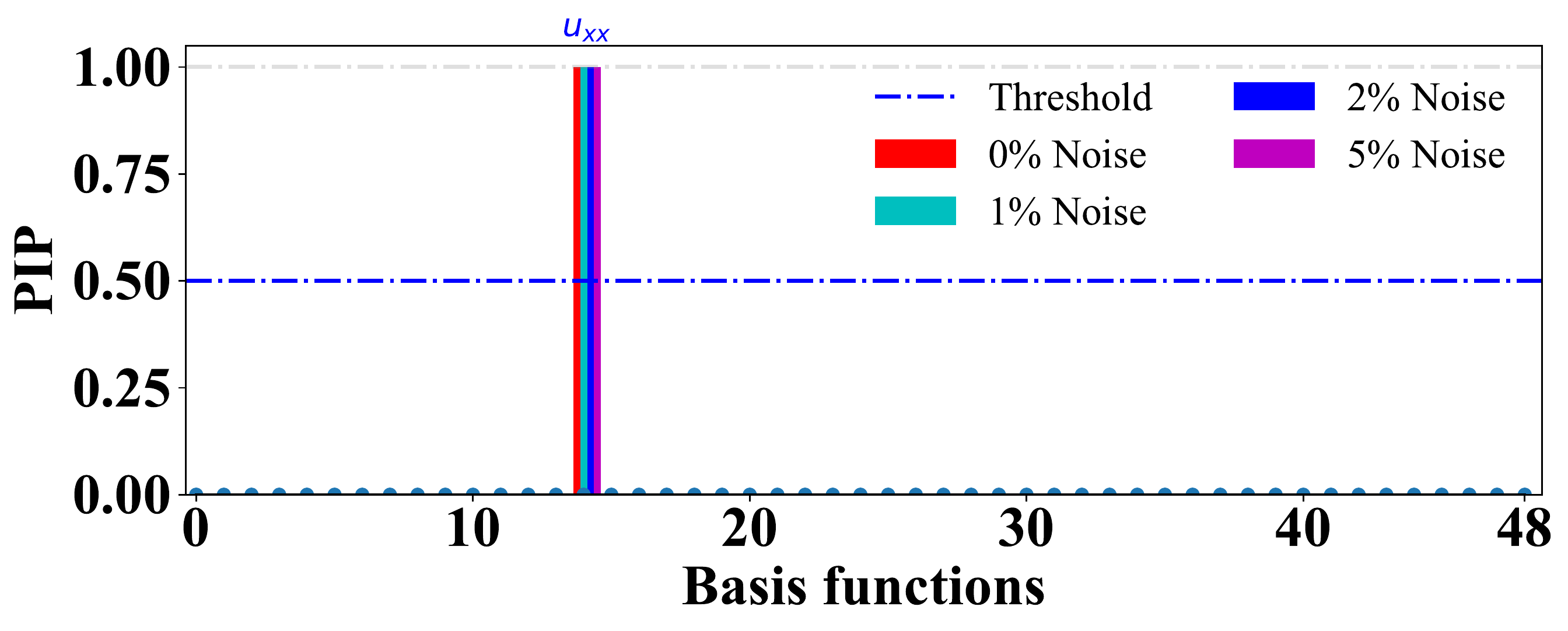}
    \caption{Identification of the basis functions for the 1D heat PDE from data corrupted with 0\%, 1\%, 2\%, and 5\% noise. Out of 49 sparse coefficients $\Phi \in \mathbb{R}^{49}$ corresponding to 49 basis functions only the basis $u_{xx}$ is found to be active. The proposed variational Bayes framework identifies the exact basis function even when the data are synthetically corrupted with 5\% zero mean Gaussian noise.}  
    \label{fig:1d_heat_basis}
\end{figure}

\begin{figure}[!ht]
    \centering
    \includegraphics[width=0.8\linewidth]{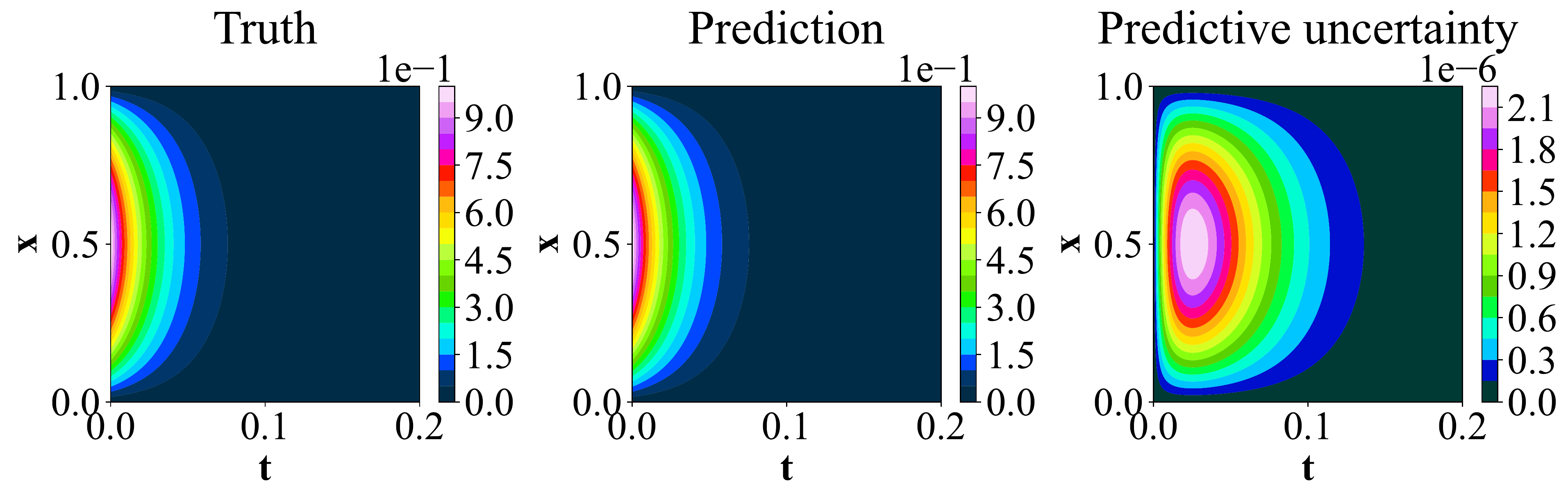}
    \caption{Predictive performance of the identified 1D heat PDE against the ground truth. Unlike the previous frequentist PDE discovery algorithms, the proposed variational Bayes framework also provides predictive uncertainty.}
    \label{fig:1d_heat_all}
\end{figure}

To discover the governing PDE for the 1-D Heat problem, a dictionary denoted as $\textbf{D} \in \mathbb{R}^{N \times 49}$ is utilized, which contains 49 basis functions. These basis functions are constructed to include derivatives up to order 6 and polynomial terms up to order 6, along with an element-wise product of $u(x, t)$. The basis functions in the dictionary are designed to capture the different possible forms of the heat equation that may arise in the data.
The selection of basis functions to represent the equation is crucial, and in this study, the model identifies the basis functions with the highest PIP value. As depicted in Fig. \ref{fig:1d_heat_basis}, the basis function picked by the model for this equation is $u_{xx}$. The identified equations are presented in Table \ref{tab:tab1}, providing information about the inferred parameters from the data analysis.
\begin{figure}
    \centering
    \includegraphics[width=0.8\textwidth]{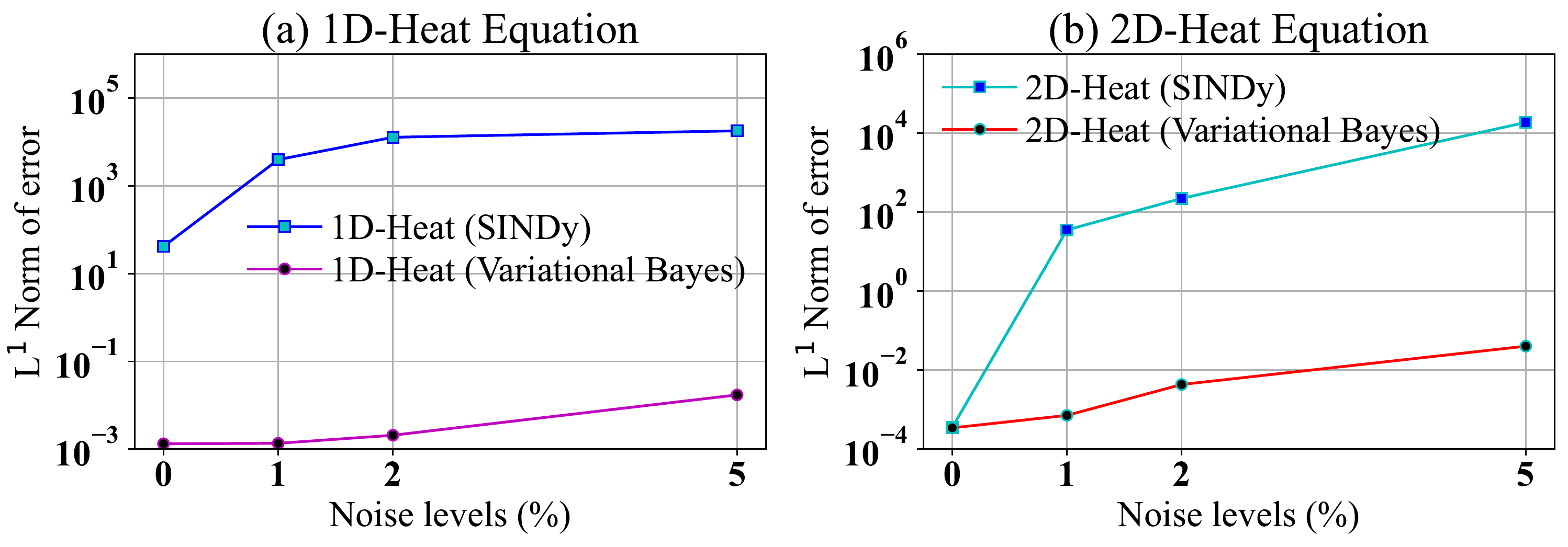}
    \caption{Identification errors for 1D and 2D heat PDE with increase in measurement noise level.}
    \label{fig:error_1d_2d_heat}
\end{figure}
We next consider a more realistic scenario where the data is corrupted by noise. To that end, the synthetic data generated previously is corrupted by up to 5\% noise. The identification errors at different noise levels are shown in Fig. \ref{fig:error_1d_2d_heat}(a). It is observed the proposed approach successfully identifies the governing physics even in the presence of noise.
Further, the response contours obtained using the identified model and true model are shown in Fig. \ref{fig:1d_heat_all}. An excellent match between the contours indicates the efficacy of the proposed approach. The predictive uncertainty estimated using the proposed approach is also shown in Fig. \ref{fig:1d_heat_all}.

\begin{table}[h!]
    \centering
        \begin{tabular}{lcc}
        \toprule
        PDEs & Correct PDE & Identified PDE \\
        \midrule
        1D Heat Eqn. & $u_t = 2.00  u_{xx}$ & $u_t = 2.001  u_{xx}$\\
        2D Heat Eqn. & $u_t = 1 u_{xx} + 1 u_{yy}$ & $u_t = 1.0002 u_{xx} + 1.0001 u_{yy}$\\
        1D Burgers Eqn. & $u_t = 1 u_{xx} - 1 u u_x$ & $u_t =0.100 u_{xx} - 1.000 u u_x$\\
        KdV Eqn.  & $u_t = -6 u u_x - u_{xxx}$ & $u_t = -5.908 u u_x - 1.05u_{xxx}$\\
        KS Eqn. & $u_t = -1uu_x -1u_{xx} - 1u_{xxxx}$  & $u_t = -0.984uu_x -0.994u_{xx} - 0.998u_{xxxx}$\\
        1D Wave Eqn. & $u_{tt} = 1 u_{xx}$  & $u_{tt} = 0.9999 u_{xx}$\\
        \bottomrule
        \end{tabular}
    \caption{Summary of the identification results of proposed variational Bayes framework for a wide class of physical systems governed by PDEs.}
    \label{tab:tab1}
\end{table}

We also considered the 2D heat equation, which in mathematical terms expressed as,
\begin{equation}\label{2d_heat}
    \begin{aligned}
        &u_t=\alpha_{1}^2 u_{xx} + \alpha_{2}^2 u_{yy}, \; x \times y \in [0,1] \times [0,1], \; t\in[0,2],\\
        &\text{IC: } u(x, y, 0)=e^{-a x^2-a y^2}, \; a=5,\\
        &\text{BC: } u_{\partial \Omega}(t) = 0.
    \end{aligned}
\end{equation}
In this equation, {$u(x, y, t)$} represents the temperature at a particular position $(x, y)$ in the two-dimensional domain at a given time $t$. The term $\alpha \in \mathbb{R}$ known as thermal diffusivity, is a positive constant that quantifies how readily heat is conducted in the material.
Similar to the previous example, we generated synthetic data by solving Eq. \eqref{2d_heat}. The spatial domain was discretized into 64 points, and a time step $\Delta t$=$0.001$s was used.

\begin{figure}[!b]
    \centering
    \includegraphics[width=0.8\textwidth]{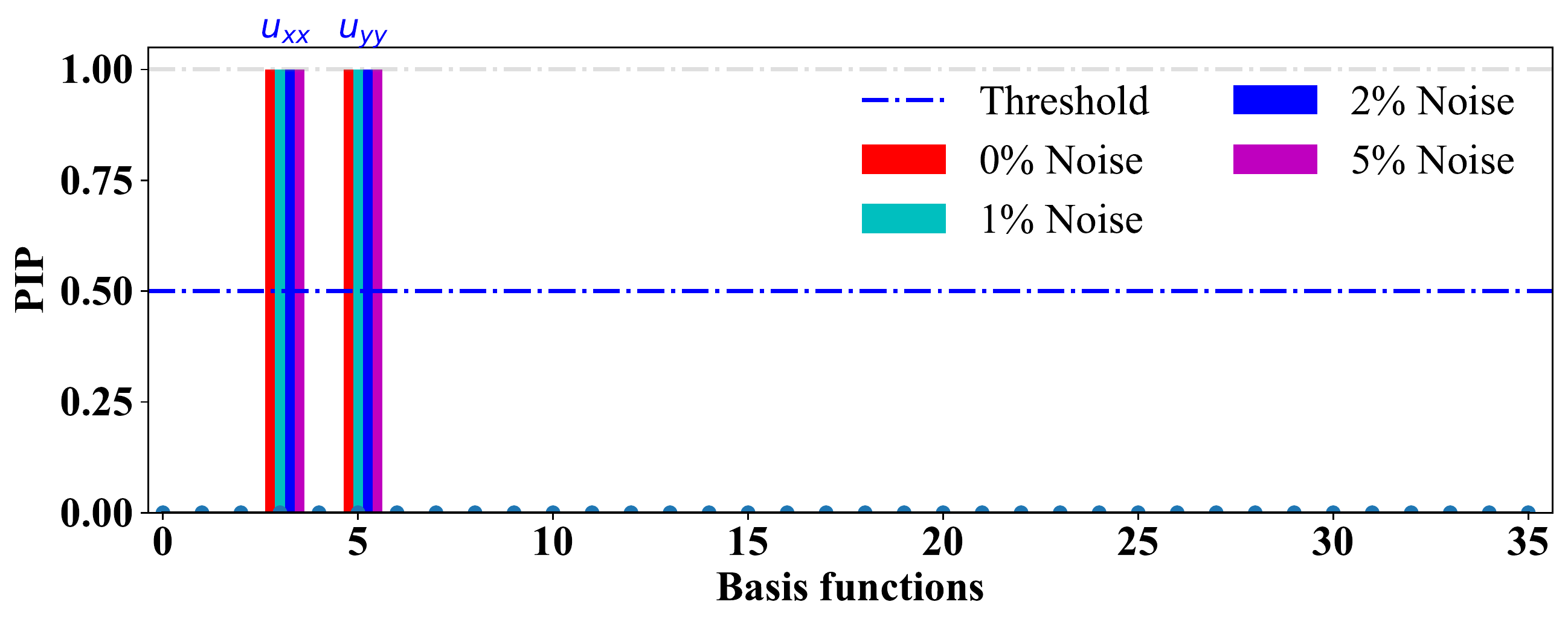}
    \caption{Identification of the basis functions for the 2D heat PDE from data corrupted with 0\%, 1\%, 2\%, and 5\% noise. Out of 35 sparse coefficients $\Phi \in \mathbb{R}^{35}$ corresponding to 35 basis functions, only the basis functions $u_{xx}$ and $u_{yy}$ are found to be active. The proposed variational Bayes framework performs consistently in all the noise levels and identifies the exact PDE model in all the cases for the 2-D heat equation.} 
    \label{fig:2d_heat_basis}
\end{figure}

\begin{figure}[!t]
    \centering
    \includegraphics[width=\textwidth]{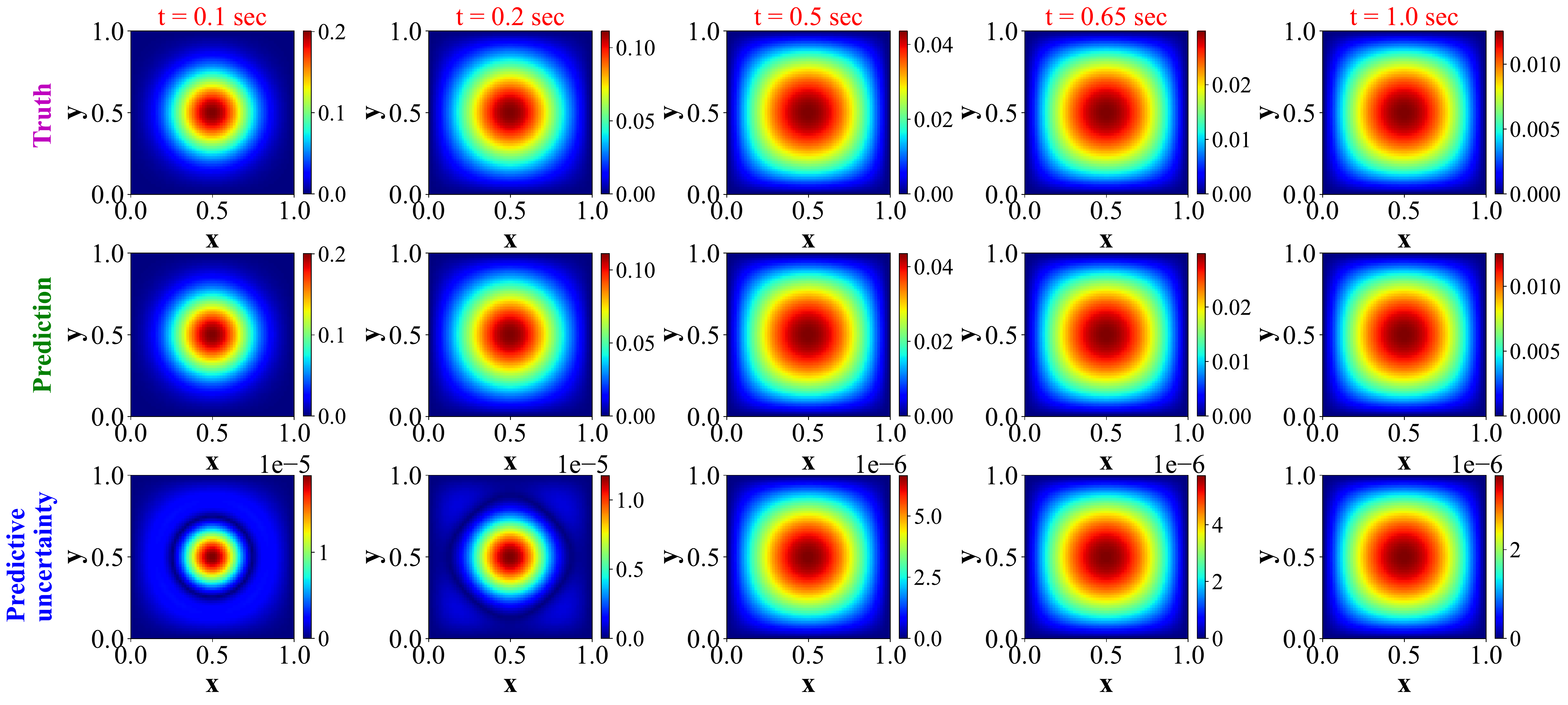}
    \caption{Predictive performance of the identified 2D heat PDE against the ground truth. The temporal evolution of the solutions of identified 2D heat PDE very closely emulates the ground truth.}
    \label{fig:2d_heat_pred}
\end{figure}

In order to identify the underlying equation from the data, a dictionary $\textbf{D} \in \mathbb{R}^{N\times35}$ of 35 basis functions is utilized. This dictionary contains derivatives up to order 2 and polynomial terms up to order 6, with an element-wise product of $u(x,t)$. The proposed approach identifies the basis functions that represent the equation. In this case, the basis functions picked by the model are $u_{xx}$ and $u_{yy}$ as shown in Fig. \ref{fig:2d_heat_basis}, indicating that these terms are required for representing the underlying heat equation (2D). Fig. \ref{fig:error_1d_2d_heat}(b) represents the error in the identified equations for different levels of noise. It is observed that the proposed approach performs satisfactorily even in the presence of noise. The response contours presented in Fig. \ref{fig:2d_heat_pred} further strengthens our claim.

\begin{figure}[!ht]
    \centering
    \includegraphics[width=0.8\textwidth]{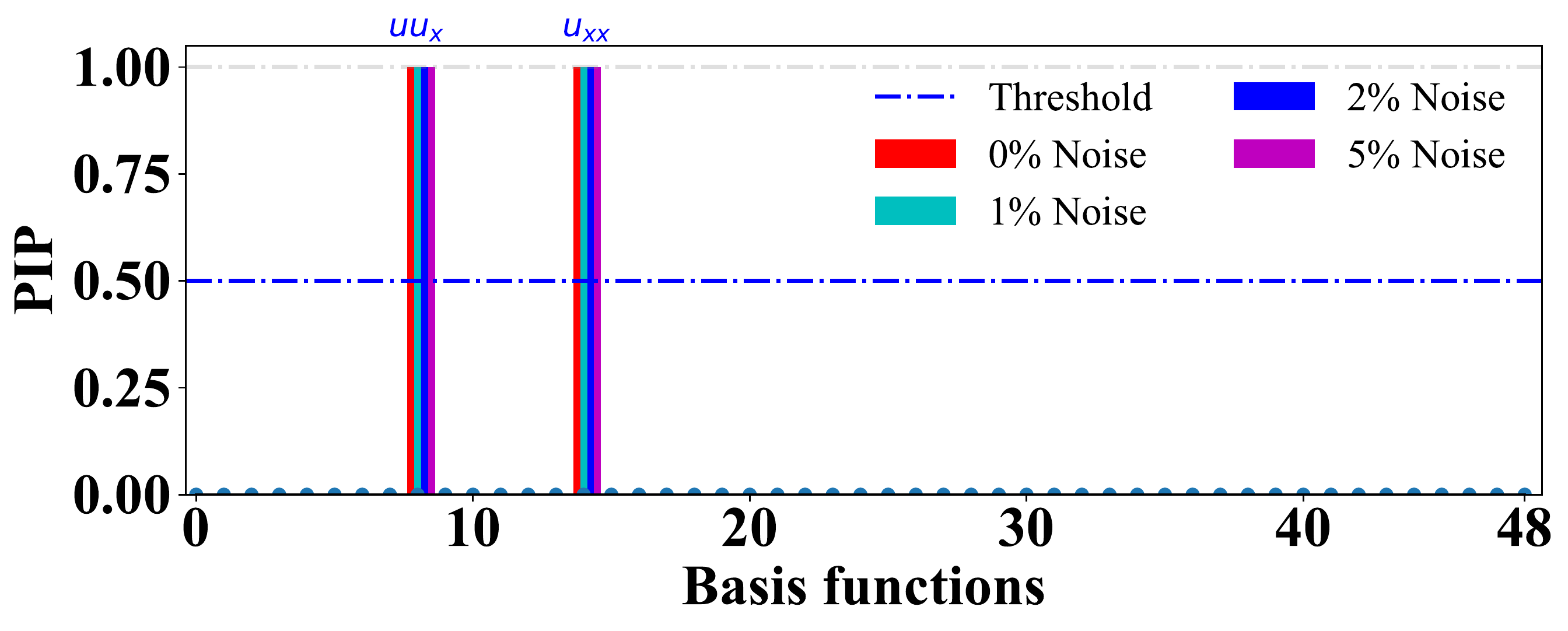}
    \caption{Identification of the basis functions for the Burgers equation from data corrupted with 0\%, 1\%, 2\%, and 5\% noise. Out of 49 sparse coefficients $\Phi \in \mathbb{R}^{49}$ corresponding to 49 basis functions the proposed variational Bayes framework exactly identifies the correct basses $uu_{x}$ and $u_{xx}$. The proposed variational Bayes framework identifies the correct basis functions in all the noise levels.}
    \label{fig:1d_burgers_basis}
\end{figure}
\subsection{Burgers equation}\label{sec:1D_Burgers}
Burgers equation is a nonlinear partial differential equation that models the behaviour of a fluid in one dimension, and is given by the following equation,
\begin{equation}\label{1d_burge_eqn}
u_t + u u_x = \nu \hspace{1mm} u_{xx}, \; x\in[-8,8], \; t\in[0,10],
\end{equation}
where $u(x,t)$ is the velocity of the fluid at position $x$ and time $t$, and $\nu \in \mathbb{R}$ is the kinematic viscosity. The first term on the left represents the change in velocity with respect to time, while the second term represents the advection of the fluid due to its own velocity. The term on the right-hand side represents the diffusion of the fluid due to its viscosity.
The nonlinear nature of Burgers equation means that it can exhibit complex phenomena, such as shock waves and turbulence. These features have made it a useful model for studying a wide range of physical systems, such as fluid flow in pipes, highway traffic flow, and gas shock waves.
We generated synthetic data by solving Eq. \eqref{1d_burge_eqn} using the finite difference method. The spatial domain was discretized into 256 points and a time step $\Delta t$=$0.09$s was used.
\begin{figure}[!b]
    \centering
    \includegraphics[width=0.8\textwidth]{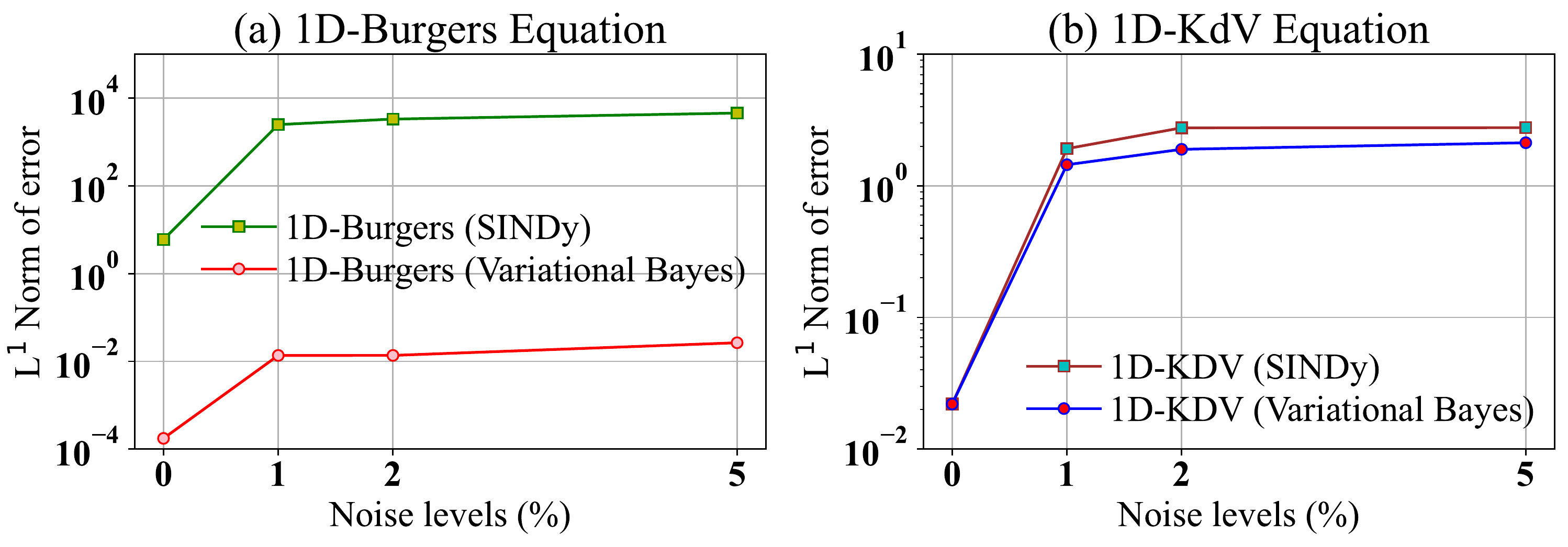}
    \caption{Identification errors for 1D Burgers' equation and Korteweg–de Vries equation with increase in measurement noise level.}
    \label{fig:error_1d_burgers_kdv}
\end{figure}

\begin{figure}[!ht]
    \centering
    \includegraphics[width=0.8\linewidth]{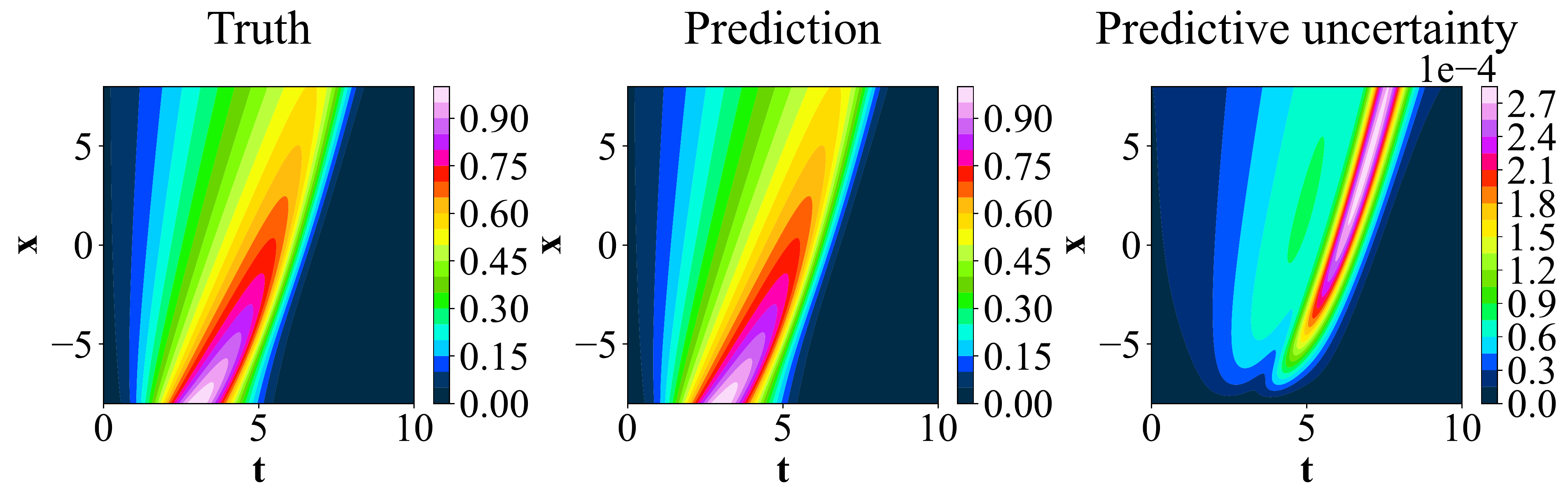}
    \caption{Predictive performance of the identified model for Burgers PDE data against the ground truth.}
    \label{fig:1d_burgers_all}
\end{figure}

In order to identify the underlying equation from the data, a dictionary $\textbf{D} \in \mathbb{R}^{N\times49}$ of 49 basis functions are utilized. This dictionary contains derivatives up to order 6 and polynomial terms up to order 6, with an element-wise product of $u(x,t)$. The proposed approach identifies the basis functions that represent the equation. In this case, the basis functions picked by the model are $u u_x$ and $u_{xx}$ as shown in Fig. \ref{fig:1d_burgers_basis}, indicating that these terms are required for representing the underlying Burgers equation. Fig. \ref{fig:error_1d_burgers_kdv}(a) represents the identification errors corresponding to different levels of noise. It is observed that the proposed approach performs satisfactorily even in the presence of noise. Further, the velocity contour presented in Fig. \ref{fig:1d_burgers_all} further strengthens the claim regarding the accuracy of the proposed approach.

\subsection{Korteweg–de Vries equation}\label{sec:Kdv}
The Korteweg-de Vries (KdV) equation is a one-dimensional nonlinear partial differential equation that describes the evolution of weakly nonlinear and long wavelength waves. The KdV equation is given by the following equation,
\begin{equation}\label{KdV_eqn}
u_t = -6 u u_x - u_{xxx}, \; x\in[-30,30], \; t\in[0, 20],
\end{equation}
where $u(x,t)$ is the dependent variable representing the wave's amplitude. 
Application of the KdV equation includes modelling of water waves, plasma waves, and Bose-Einstein condensates.
Similar to the previous example, we generated synthetic data by solving Eq. \eqref{KdV_eqn} using the finite difference method. The spatial domain was discretized into 512 points and a time step $\Delta t$ = $0.09$s was used.

\begin{figure}[!ht]
    \centering
    \includegraphics[width=0.8\textwidth]{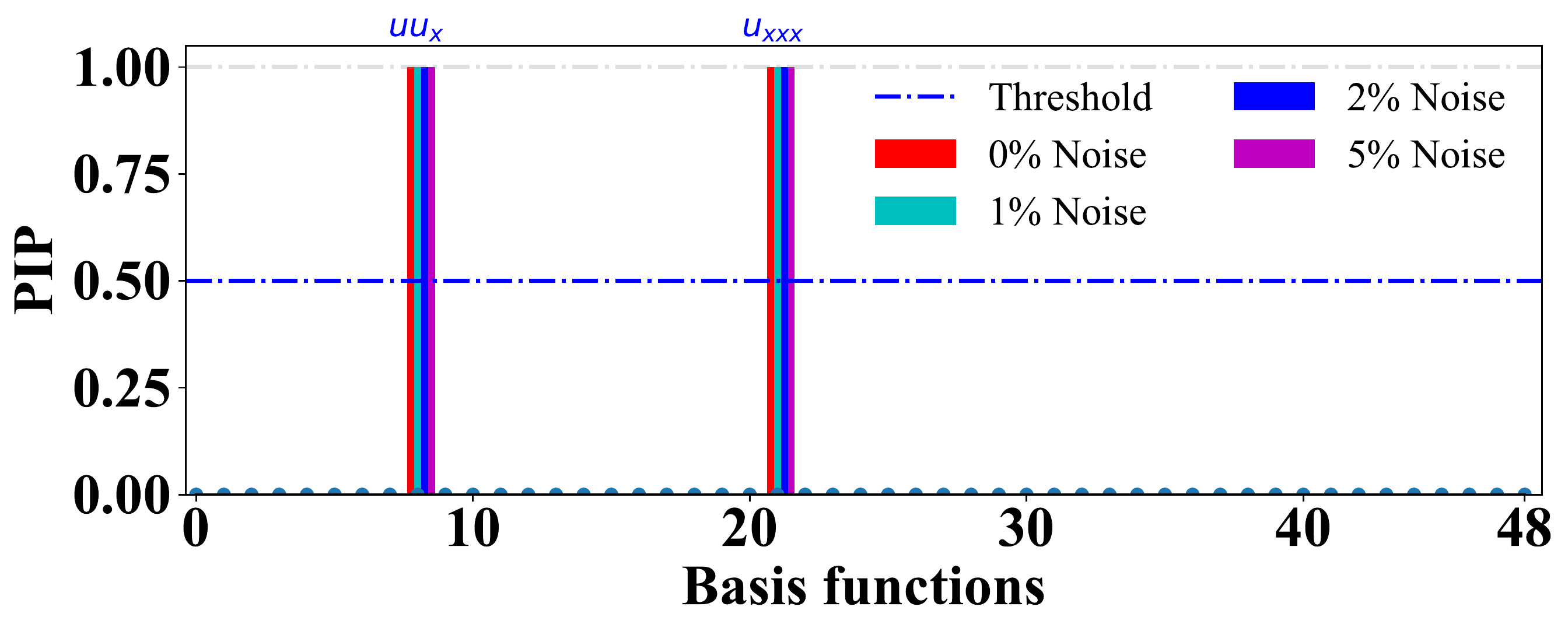}
    \caption{Identification of the basis functions for the third order Korteweg–de Vries equation from data corrupted with 0\%, 1\%, 2\%, and 5\% noise. Out of 49 sparse coefficients $\Phi \in \mathbb{R}^{49}$ corresponding to 49 basis functions the proposed variational Bayes framework exactly identifies the correct basses $uu_{x}$ and $u_{xxx}$.}  
    \label{fig:kdv_basis}
\end{figure}

\begin{figure}[!t]
    \centering
    \includegraphics[width=0.8\linewidth]{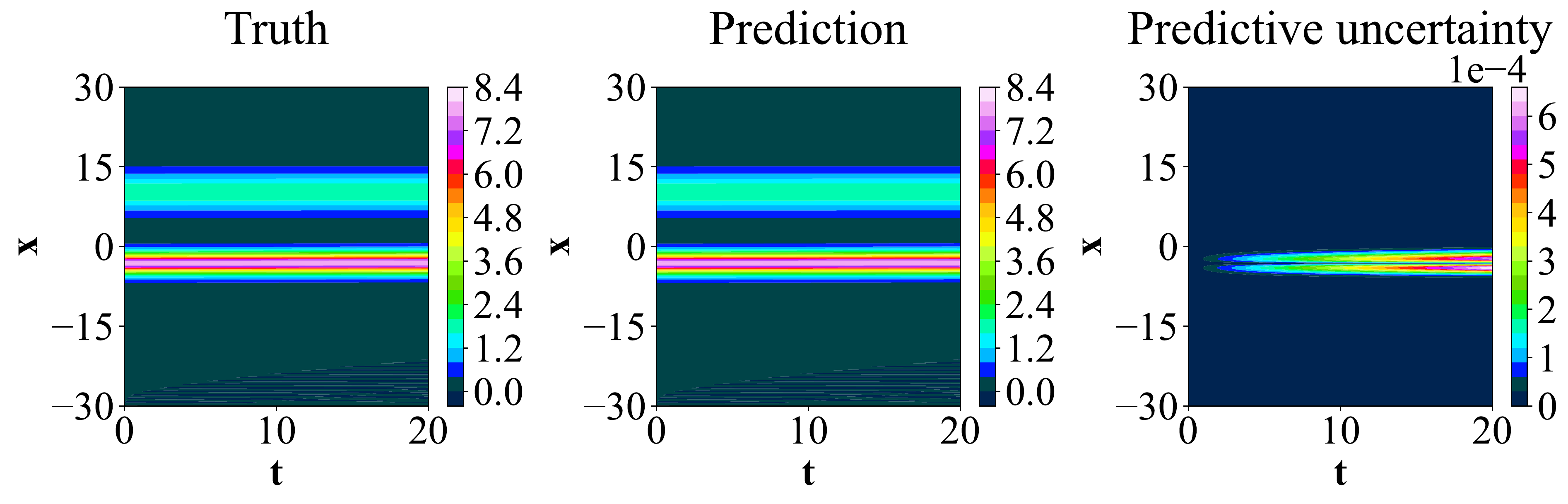}
    \caption{Predictive performance of the identified model for KdV PDE data against the ground truth.}
    \label{fig:kdv_all}
\end{figure}

In order to identify the underlying equation from the data, the dictionary $\textbf{D}\in\mathbb{R}^{N\times49}$ of 49 basis functions is used, which contains derivatives up to order 6 and polynomial terms up to order 6, along with an element-wise product of $u(x, t)$.
The proposed approach identifies the basis functions that represent the equation. In this case, the basis functions picked by the model are $u u_x$ and $u_{xxx}$ as shown in Fig. \ref{fig:kdv_basis}, indicating that these terms are required for representing the underlying KdV equation. Fig. \ref{fig:error_1d_burgers_kdv}(b) portrays the identification errors corresponding to different levels of noise. Although the basis functions are identified correctly,  the parameters of the identified equation deviate from the ground truth as the noise level increases; this is primarily because of the presence of a third-order derivative present in the governing physics. The contours shown in Fig. \ref{fig:kdv_all} further confirm the efficacy of the proposed approach.

\subsection{Kuramoto Sivashinsky equation}\label{sec:KuraSiva}
The Kuramoto-Sivashinsky (KS) equation is a partial differential equation that describes the spatio-temporal evolution of a thin film of viscous fluid or a one-dimensional flame front and is given as,
\begin{equation}\label{KS_eqn}
u_t = -uu_x -u_{xx} - u_{xxxx}, \; x\in[0,100], \; t\in[0,100],            
\end{equation}
where $u(x,t)$ is the height of the fluid film or the temperature of the flame front at position $x$ and time $t$, the equation is nonlinear, fourth-order, and exhibits chaotic behaviour, making it an important model system for studying spatiotemporal chaos and pattern formation. 
Similar to the previous example, we generated synthetic data by solving Eq. \eqref{KS_eqn} using the finite difference method. The spatial domain was discretized into 1024 points and a time step $\Delta t$=$0.4$s was used.
\begin{figure}[!ht]
    \centering
    \includegraphics[width=0.8\textwidth]{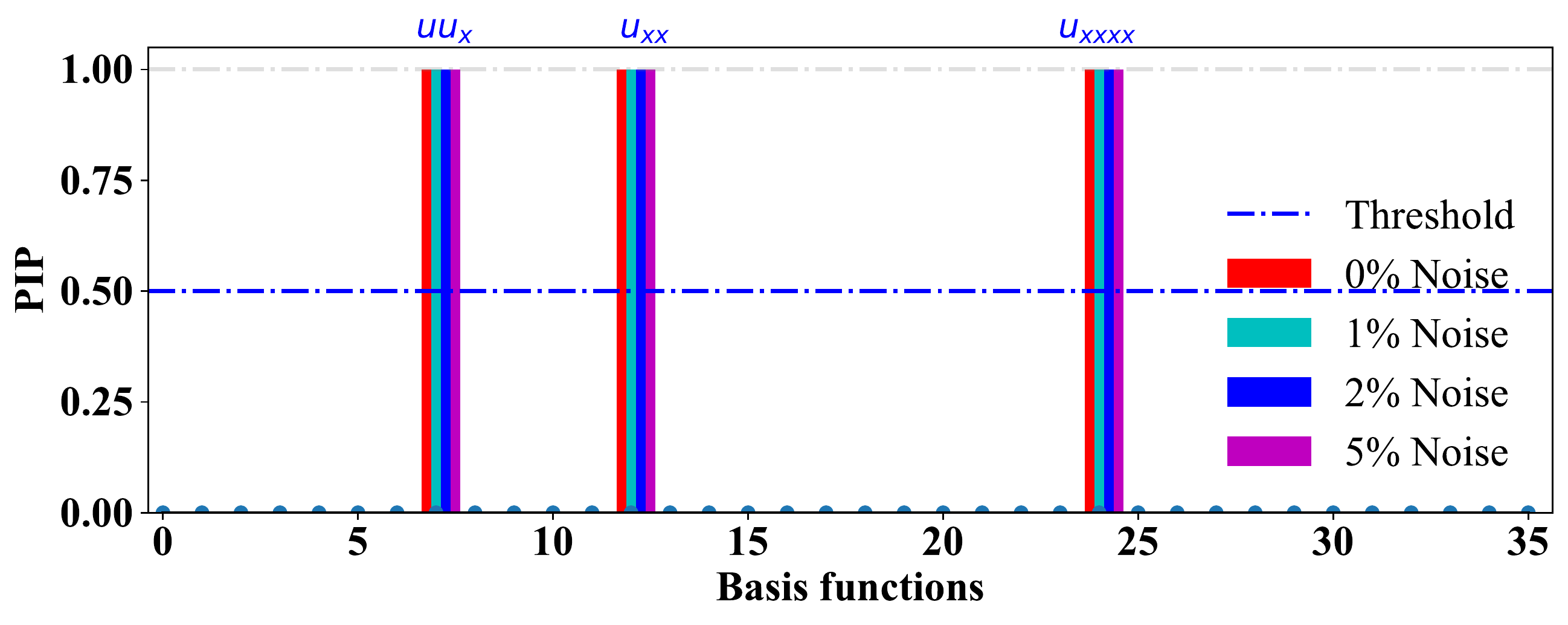}
    \caption{Identification of the basis functions for the fourth order chaotic Kuramoto Sivashinsky equation from data corrupted with 0\%, 1\%, 2\%, and 5\% noise. A total of 36 basis functions are used in the dictionary and out of 36 sparse coefficients $\Phi \in \mathbb{R}^{36}$ corresponding to 36 basis functions the proposed variational Bayes framework exactly identifies the correct basses $uu_{x}$, $u_{xx}$, and $u_{xxx}$. At all the noise levels, the proposed variational Bayes framework performs consistently and identifies the correct basis functions. Even at the 5\% noise level, the proposed framework accurately identifies the fourth-order basis function, which is generally very noise sensitive.} 
    \label{fig:kurasiva_basis}
\end{figure}

In order to identify the underlying equation from the data, the dictionary $\textbf{D}\in\mathbb{R}^{N\times35}$ of 35 basis functions is used, which contains derivatives up to order 5 and polynomial terms up to order 5, along with an element-wise product of $u(x, t)$.
The proposed approach identifies the basis functions that represent the equation. In this case, the basis functions picked by the model are $u_x, u_{xx}$, and $u_{xxxx}$ as shown in Fig. \ref{fig:kurasiva_basis}, indicating that these terms are required for representing the underlying KS equation. Fig. \ref{fig:error_ks_wave}(a) displays the error in the identification for different levels of noise. The proposed approach successfully identifies the correct basis functions; however, the parameters deviate in the presence of noise. This is because of the fourth-order derivative present in the equation.

\subsection{1-D Wave equation}\label{sec:1D_Wave}
\begin{figure}
    \centering
    \includegraphics[width=0.8\linewidth]{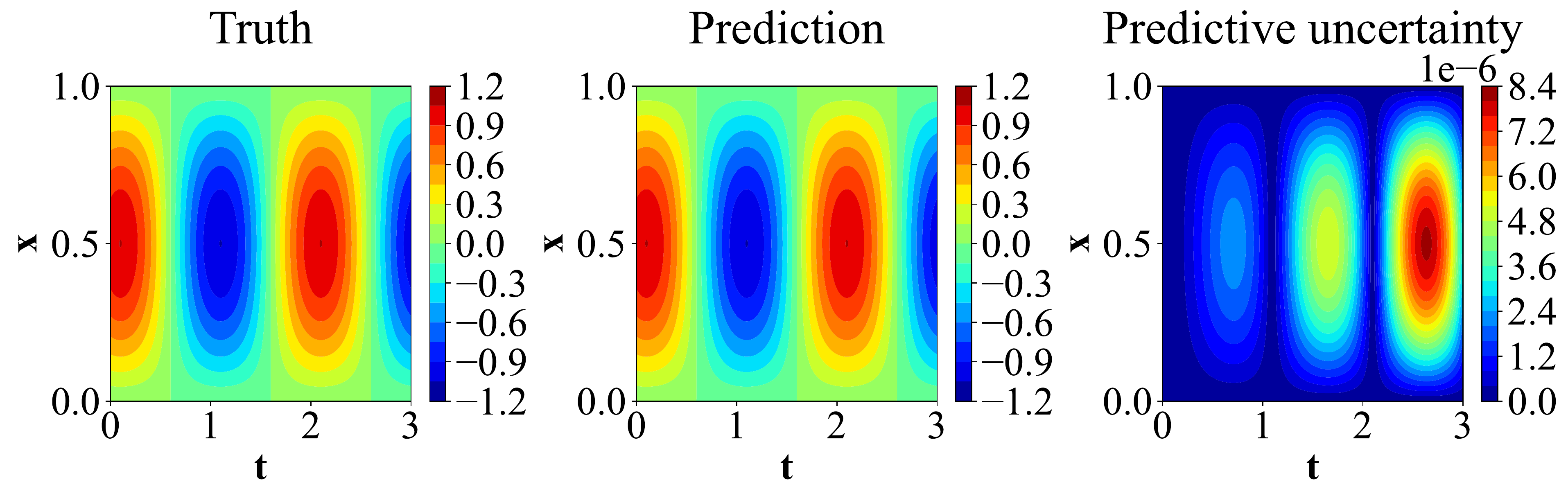}
    \caption{Predictive performance of the identified PDE model for Wave equation against the ground truth.}
    \label{fig:wave_all}
\end{figure}

The one-dimensional wave equation is a mathematical model that describes the behaviour of waves in one dimension, such as waves on a string or sound waves in a pipe. It is a partial differential equation that relates the second time derivative of the wave function to the second spatial derivative of the wave function and is given by,

\begin{equation}\label{1d_wave_eqn}
    \begin{aligned}
        & u_{tt} = \alpha^2 u_{xx}, \; x\in[0,1], \; t\in[0,3],\\
        &\text{IC: } u(x,0) = sin(\pi x),\\
        &\text{BC: } u(0,t) = 0, \; u(1,t) = 0,
    \end{aligned}
\end{equation}
where $u(x, t)$ represents the wave displacement at position $x$ and time $t$, and $\alpha$ represents the wave speed. Many important applications of the wave equation can be found in engineering, physics, and mathematics. It is used to analyze and design structures like bridges and buildings. It is also used to study basic physical phenomena, of the behavior of light and sound. Its solutions have a wide range of practical applications, including musical instrument design, communication system optimization, and earthquake and other natural disaster detection.
Similar to the previous example, we generated synthetic data \ref{fig:wave_all} by solving Eq. \ref{1d_wave_eqn} using the finite difference method. The spatial domain was discretized into 100 points and a time step $\Delta s$=$0.003$s was used.
\begin{figure}[!ht]
    \centering
    \includegraphics[width=0.8\textwidth]{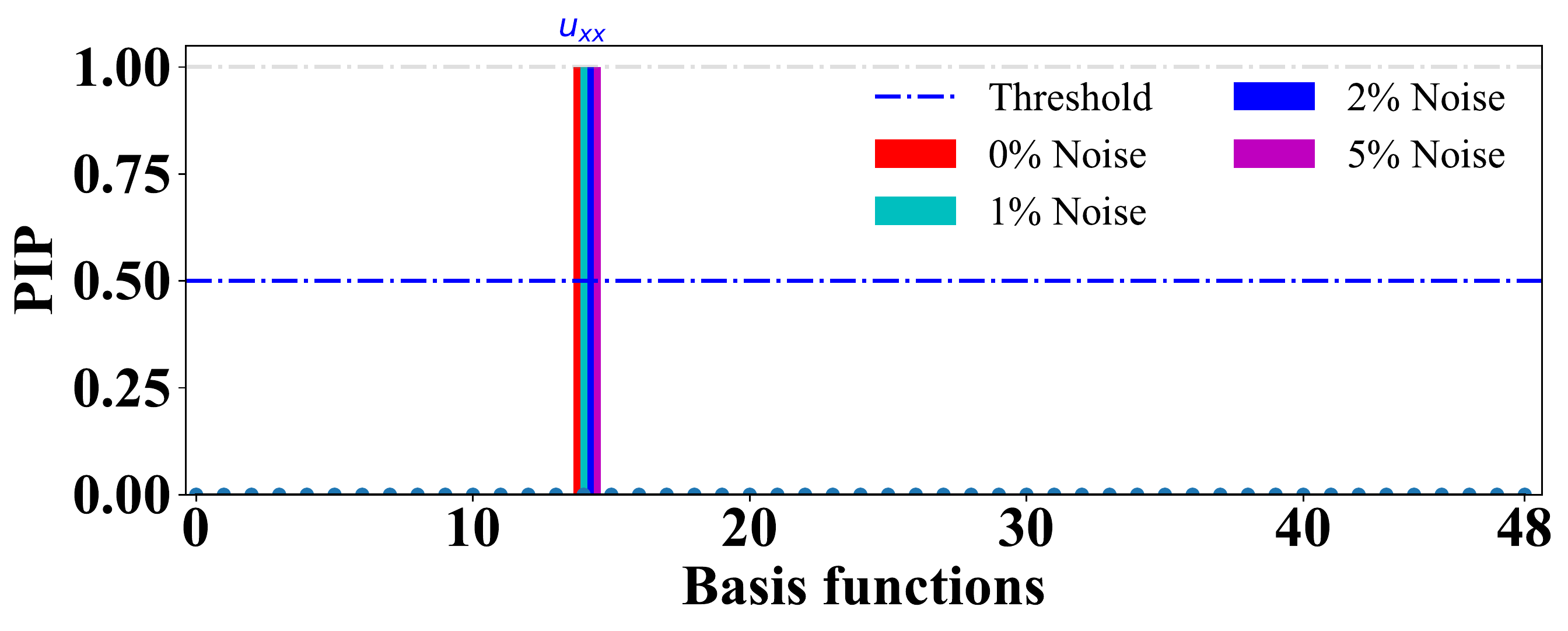}
    \caption{Identification of the basis functions for the 1D Wave equation from data corrupted with 0\%, 1\%, 2\%, and 5\% noise. A total of 49 basis functions are utilized and out of the 49 sparse coefficients $\Phi \in \mathbb{R}^{49}$ corresponding to 49 basis functions the proposed variational Bayes framework exactly identifies the correct basis $u_{xx}$. The performance across all the noise levels is found consistent.} 
    \label{fig:1d_wave_basis}
\end{figure}

\begin{figure}[!ht]
    \centering
    \includegraphics[width=0.8\textwidth]{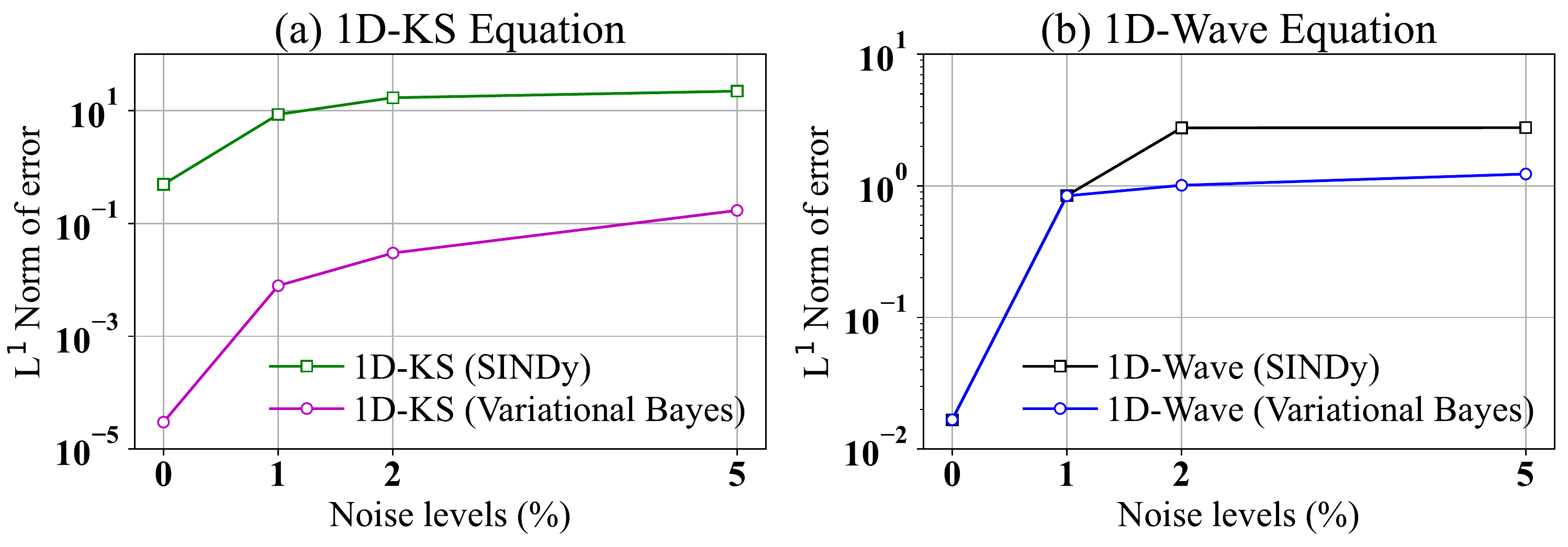}
    \caption{Identification errors for Kuramoto Sivashinsky PDE and 1D wave PDE with increase in measurement noise level.}
    \label{fig:error_ks_wave}
\end{figure}

In order to identify the underlying equation from the data, the dictionary $\textbf{D}\in\mathbb{R}^{N\times49}$ of 49 basis functions is used, which contains derivatives up to order 6 and polynomial terms up to order 6, along with an element-wise product of $u(x, t)$.
The proposed approach identifies the basis functions that represent the equation. In this case, the basis functions picked by the model is $u_{xx}$ as shown in Fig. \ref{fig:1d_wave_basis}, indicating that these terms are required for representing the underlying wave equation. Fig. \ref{fig:error_ks_wave}(b) shows the evolution of identification error with different levels of noise. It is observed that the proposed approach performs satisfactorily even in the presence of noise. The contours shown in Fig. \ref{fig:wave_all} further enforce the claim regarding the accuracy of the result.

\subsection{Comparison between SINDy algorithm and proposed approach}\label{Comparison}
In addition to the comparisons in Figs. \ref{fig:error_1d_2d_heat}, \ref{fig:error_1d_burgers_kdv}, and \ref{fig:error_ks_wave}, we have further performed a comparison between the SINDy algorithm and the proposed variational Bayes framework in terms of false positive rate, as shown in Table \ref{tab:Comparison}. The false positive rate provides information about the number of candidate functions that do not accurately represent the underlying equation.
Similar to the previous study, this comparison is carried out at four different noise levels (i.e., 0\%, 1\%, 2\%, and 5\%). At all the noise levels, the proposed framework consistently identifies the exact basis functions of the actual model. On the other hand, the existing SINDy algorithm identifies the exact basses for only Kuramoto Sivashinsky and the 2D heat equation. The performance of the SINDy algorithm further deteriorates with an increase in noise levels. Therefore these results are a clear indication of the robustness of the proposed framework against low noise levels up to 5\% of the standard deviation of the actual signal.

\begin{table}[ht!]
    \centering
    \small
    \caption{Comparison between the identification results of proposed variational Bayes approach and existing SINDy algorithm.}
    \begin{tabular}{llllll}
    \toprule
    \multirow{2}{*}{Example} & \multirow{2}{*}{Algorithms} & \multicolumn{4}{c}{False Positive rate} \\
    \cmidrule(r){3-6}
     && Noise = 0\% & Noise = 1\% & Noise = 2\% & Noise = 5\% \\ 
    \midrule
    1-D Heat Eqn. & SINDy & 0.0204 & 0.1224 & 0.3061 & 0.1020 \\
    & Variational Bayes & 0 & 0 & 0 & 0 \\
    \midrule
    Burgers Eqn. & SINDy & 0.1428 & 0.1836 & 0.0816 & 0.1224 \\
    & Variational Bayes & 0 & 0 & 0 & 0 \\
    \midrule
    Korteweg–de Vries Eqn. & SINDy & 0.2040 & 0.5918 & 0.3265 & 0.5510 \\
    & Variational Bayes & 0 & 0 & 0 & 0 \\
    \midrule
    Kuramoto Sivashinsky Eqn. & SINDy & 0 & 0.5714 & 1 & 0.8571 \\
    & Variational Bayes & 0 & 0 & 0 & 0 \\
    \midrule
    1-D Wave Eqn. & SINDy & 0.0204 & 0.0204 & 0.1836 & 0.1632 \\
    & Variational Bayes & 0 & 0 & 0 & 0 \\
    \midrule
    2-D Heat Eqn. & SINDy & 0 & 0.3142 & 0.6149 & 0.7714 \\
    & Variational Bayes & 0 & 0 & 0 & 0 \\
    \bottomrule
    \end{tabular}
    \label{tab:Comparison}
\end{table}

\section{Conclusion}\label{conclusion}
In scientific machine learning, the task of identifying PDE accurately from noisy and extensive data poses a significant challenge. In this paper, we propose a sparse Bayesian learning algorithm for identifying governing PDE from data. By leveraging a sparsity-promoting spike-and-slab prior, we are able to selectively choose the most relevant functions from a manually-designed dictionary of candidate functions. As a result, we transform the problem into a sparse linear regression task and employ variational Bayes to solve the same, which has proven to be more effective than the widely-used SINDy method when handling noisy data. The proposed algorithm employing variational Bayes addresses the computational cost associated with sampling-based approaches.
To ensure interpretability in our model, we construct a dictionary of candidate functions manually and enforce sparsity in the corresponding parameters. These candidate functions mainly consist of spatial and temporal derivatives of the solution. We utilize finite difference methods for taking derivatives when the data is free from noise, while in the presence of noise, we employ polynomial interpolation methods. 

However, it is essential to note that data points situated near the boundaries, where polynomial fitting becomes challenging, should be avoided as they can lead to inaccurate approximations.
One of the main challenges associated with this method is that the accuracy of the results heavily depends on the degree of the polynomial and the number of data points used to fit it. Therefore, it is crucial to meticulously select the degree of the polynomial and the number of points to ensure that the results are as accurate as possible.
Our sparsity-based approach guarantees that only the most significant features of the PDE are retained in the model, leading to a more interpretable solution. The proposed approach identifies the candidate function that exhibits a high correlation with the true candidate, ensuring accurate model identification.
To evaluate the effectiveness of the VB algorithm, we apply it to five example problems: the heat equation (1D and 2D), Burgers equation, KdV equation, Wave equation, and Kuramoto-Sivashinsky equation. Our results demonstrate the algorithm's efficacy and robustness, as we successfully identify the correct models for all cases. However, we do observe that error propagation is more pronounced when taking derivatives of noisy data, particularly in the case of the Kuramoto-Sivashinsky equation. This limitation can be addressed by incorporating denoising techniques into the process.

In summary, our proposed approach provides an interpretable solution for identifying PDEs, making it a valuable tool for various applications in science and engineering. By combining the power of Bayesian methods with a sparse regression approach, we have showcased the effectiveness of the VB algorithm in accurately identifying the model structure and determining its parameters for PDEs. This approach has the potential to enhance our understanding of complex systems and improve our ability to make accurate predictions across a wide range of fields.

\section*{Acknowledgements} 
T. Tripura acknowledges the financial support received from the Ministry of Education (MoE), India, in the form of the Prime Minister's Research Fellowship (PMRF). S. Chakraborty acknowledges the financial support received from Science and Engineering Research Board (SERB) via grant no. SRG/2021/000467, Ministry of Port and Shipping via letter no. ST-14011/74/MT (356529), and seed grant received from IIT Delhi. R. Nayek acknowledges the financial support received from Science and Engineering Research Board (SERB) via grant no. SRG/2022/001410, Ministry of Port and Shipping via letter no. ST-14011/74/MT (356529), and seed grant received from IIT Delhi.

\section*{Declarations}


\subsection*{Conflicts of interest} The authors declare that they have no conflict of interest.


\subsection*{Code availability} Upon acceptance, all the source codes to reproduce the results in this study will be made available to the public on GitHub by the corresponding author.

\appendix

\section{Variational Bayesian inference for variable selection}\label{appendixA}
To compute the SS priors, we need to obtain the posterior distribution $p\left({\Phi}, \boldsymbol{Z}, \sigma^{2} \mid \boldsymbol{Y}\right)$ in Eq. \eqref{eq:Bayes} of the model parameters.
However, obtaining the posterior distribution analytically is often impossible due to the presence of a multi-dimensional intractable integral involving the marginal likelihood $p(\boldsymbol{Y})$ of the data. 
To overcome this challenge, MCMC-based methods can be used. Although MCMC-based methods can provide fairly accurate results, they can be computationally expensive, especially when dealing with high-dimensional data and complex models.

The variational Bayes (VB) approach is used in this study to approximate the joint posterior distribution $p\left({\Phi}, \boldsymbol{Z}, \sigma^{2} \mid \boldsymbol{Y}\right)$ using simpler variational distributions. The presence of a Dirac-delta function in the SS prior makes closed-form derivatives of the VB approach difficult. As a result, the linear regression model with SS prior must be reparameterized in a way that is more suitable for the variational Bayes \cite{nayek2021spike} approach. In order to accomplish this, the SS prior is reformulated as,
\begin{equation}
    \boldsymbol{Y} \mid {\Phi}, \boldsymbol{Z}, \sigma^{2} \sim \mathcal{N}\left(\mathbf{D} \boldsymbol{\Lambda}{\Phi}, \sigma^{2} \mathbf{I}_{N}\right),
    \label{eq:ss_prior}
\end{equation}
with, $\sigma^{2} \sim \mathcal{I} \mathcal{G}\left(a_{\sigma}, b_{\sigma}\right)$,
    $\phi_{i} \sim \mathcal{N}\left(0, \sigma^{2} v_{s}\right)$, $Z_{i} \sim \operatorname{Bern}\left(p_{0}\right)$, $i=1, \ldots, K$, and $\boldsymbol{\Lambda}=\operatorname{diag}\left(Z_{1}, \ldots, Z_{K}\right)$. The goal of variational Bayes is to approximate the true posterior distribution $p\left({\Phi}, \boldsymbol{Z}, \sigma^{2} \mid \boldsymbol{Y}\right)$ with a variational distribution $q({\Phi}, \boldsymbol{Z}, \sigma^{2})$ that belongs to a tractable family of distributions $Q$, such as the Gaussian distribution. The best approximation $q^*\in Q$ is then determined by minimizing the Kullback-Leibler (KL) divergence \cite{joyce2011kullback} between $q({\Phi}, \boldsymbol{Z}, \sigma^{2})$ and $p\left({\Phi}, \boldsymbol{Z}, \sigma^{2} \mid \boldsymbol{Y}\right)$, i.e., (see Ref. \cite{nayek2022equation}),
\begin{align}
    q^{*}\left({\Phi}, \boldsymbol{Z}, \sigma^{2}\right)
    &=\underset{q \in Q}{\arg \min }  \operatorname{KL}\left[q\left({\Phi}, \boldsymbol{Z}, \sigma^{2}\right) \| p\left({\Phi}, \boldsymbol{Z}, \sigma^{2} \mid \boldsymbol{Y}\right)\right],\label{eq:KL_div}\\
    &=\underset{q \in Q}{\arg \min } \mathbb{E}_{q\left({\Phi}, \boldsymbol{Z}, \sigma^{2}\right)}\left[\ln \left(\frac{q\left({\Phi}, \boldsymbol{Z}, \sigma^{2}\right)}{p\left({\Phi}, \boldsymbol{Z}, \sigma^{2} \mid \boldsymbol{Y}\right)}\right)\right]\label{eq:KL_Div_2}.
\end{align}
In the Eq. \eqref{eq:KL_div}, the notation $\mathbb{E}_{q\left({\Phi}, \boldsymbol{Z}, \sigma^{2}\right)}[\cdot]$ represents the expectation taken with respect to the variational distribution $q\left({\Phi}, \boldsymbol{Z}, \sigma^{2}\right)$. By expanding the KL divergence term in the Eq. \eqref{eq:KL_Div_2}, we arrive at an expression for the evidence lower bound (ELBO), 
\begin{equation}
    \begin{aligned}
    \mathrm{KL}\left[q\left({\Phi}, \boldsymbol{Z}, \label{KL}\sigma^{2}\right) \| p\left({\Phi}, \boldsymbol{Z}, \sigma^{2} \mid \boldsymbol{Y}\right)\right] &=\mathbb{E}_{q\left({\Phi}, \boldsymbol{Z}, \sigma^{2}\right)}\left[\ln \left(\frac{q\left({\Phi}, \boldsymbol{Z}, \sigma^{2}\right)}{p\left({\Phi}, \boldsymbol{Z}, \sigma^{2} \mid \boldsymbol{Y}\right)}\right)\right], \\
    &=\ln p(\boldsymbol{Y})-\mathrm{ELBO}.
    \end{aligned}
\end{equation}
Here, $\ln{p(\boldsymbol{Y})}$ is a constant term with respect to the variational distribution $q\left({\Phi}, \boldsymbol{Z}, \sigma^{2}\right)$. Note KL divergence is a positive quantity; therefore ELBO is the lower bound of $\ln{p(\boldsymbol{Y})}$, which implies minimizing the KL divergence is equivalent to maximizing the ELBO,
\begin{subequations}
    \begin{equation}
        {\operatorname{KL}\left[q\left({\Phi}, \boldsymbol{Z}, \sigma^{2}\right) \| p\left({\Phi}, \boldsymbol{Z}, \sigma^{2}\right)\right]-\mathbb{E}_{q\left({\Phi}, \boldsymbol{Z}, \sigma^{2}\right)}\left[\ln p\left(\boldsymbol{Y} \mid {\Phi}, \boldsymbol{Z}, \sigma^{2}\right)\right]}  = -\text{ELBO}. 
    \end{equation}
    \begin{equation} \label{eq:ELBO}
        q^{*}\left({\Phi}, \boldsymbol{Z}, \sigma^{2}\right)=\underset{q \in Q}{\arg \max } \underbrace{\mathbb{E}_{q\left({\Phi}, \boldsymbol{Z}, \sigma^{2}\right)}\left[\ln p\left(\boldsymbol{Y} \mid {\Phi}, \boldsymbol{Z}, \sigma^{2}\right)\right]-\mathrm{KL}\left[q\left({\Phi}, \boldsymbol{Z}, \sigma^{2}\right) \| p\left({\Phi}, \boldsymbol{Z}, \sigma^{2}\right)\right]}_{\text {ELBO }}.
    \end{equation}
\end{subequations}
Let $q({\Phi}, \boldsymbol{Z}, \sigma^2)$ be a factorized distribution, where each parameter group is assigned a separate variational distribution. Specifically,
\begin{equation}
    q\left({\Phi}, \boldsymbol{Z}, \sigma^{2}\right)=q({\Phi}) q\left(\sigma^{2}\right) \prod_{i=1}^{K} q\left(Z_{i}\right),
\end{equation}
where $q({\Phi})$ is a multivariate normal distribution over the parameters ${\Phi}$, i.e., $q({\Phi}) =\mathcal{N}\left(\boldsymbol{\mu}^{q}, \boldsymbol{\Sigma}^{q}\right)$, $q(\sigma^2)$ is an inverse Gamma distribution over the noise variance $\sigma^2$, i.e., $q\left(\sigma^{2}\right) =\mathcal{I} G\left(a_{\sigma}^{q}, b_{\sigma}^{q}\right)$, and $q(Z_i)$ is a Bernoulli distribution over each binary indicator variable $Z_i$, i.e., $q\left(Z_{i}\right) =\operatorname{Bern}\left(w_{i}^{q}\right)$ that determines whether the $i$-th component of the mixture model is included or not.

The variables $\boldsymbol{\mu}^{q}$, $\boldsymbol{\Sigma}^{q}$, $a_{\sigma}^{q}$, and $b_{\sigma}^{q}$ are the parameters of individual variational distributions. $w_{i}^{q}$ is the probability that the $i$-th basis from the dictionary is included. These parameters are learned by optimizing the ELBO.
The relationships that lead to the optimal selection of variational parameters by maximizing the evidence lower bound (ELBO) in Eq.\ \eqref{eq:ELBO} are as follows,
\begin{subequations}\label{eq:var_par1}
    \begin{align}
    q^{*}({\Phi}) & \propto \mathbb{E}_{q(\boldsymbol{Z}) q\left(\sigma^{2}\right)}\left[\ln p\left(\boldsymbol{Y}, {\Phi}, \boldsymbol{Z}, \sigma^{2}\right)\right], \\
    q^{*}(\boldsymbol{Z}) & \propto \mathbb{E}_{q({\Phi}) q\left(\sigma^{2}\right)}\left[\ln p\left(\boldsymbol{Y}, {\Phi}, \boldsymbol{Z}, \sigma^{2}\right)\right], \\
    q^{*}(\sigma^{2}) & \propto \mathbb{E}_{q({\Phi}) q(\boldsymbol{Z})}\left[\ln p\left(\boldsymbol{Y}, {\Phi}, \boldsymbol{Z}, \sigma^{2}\right)\right].
    \end{align}
\end{subequations}
The variational parameters are obtained \eqref{eq:variationa_param} after solving the above equations. For a more comprehensive and detailed understanding, please refer to: \cite{nayek2022equation}:
\begin{subequations}\label{eq:variationa_param}
    \begin{align}
        \boldsymbol{\Sigma}^{q} &=\left[\tau\left(\left(\mathbf{D}^{T} \mathbf{D}\right) \odot \boldsymbol{\Omega}+v_{s}^{-1} \mathbf{I}_{K}\right)\right]^{-1}, \\
        \boldsymbol{\mu}^{q} &=\tau \boldsymbol{\Sigma}^{q} \mathbf{W}^{q} \mathbf{D}^{T} \boldsymbol{Y}, \\
        a_{\sigma}^{q} &=a_{\sigma}+0.5 N+0.5 K, \\
        b_{\sigma}^{q} &=b_{\sigma}+0.5\left[\boldsymbol{Y}^{T} \boldsymbol{Y}-2 \boldsymbol{Y}^{T} \mathbf{D W}^{q} \boldsymbol{\mu}^{q}+\operatorname{tr}\left\{\left(\left(\mathbf{D}^{T} \mathbf{D}\right) \odot \boldsymbol{\Omega}+v_{s}^{-1} \mathbf{I}_{K}\right)\left(\boldsymbol{\mu}^{q} \boldsymbol{\mu}^{q T}+\boldsymbol{\Sigma}^{q}\right)\right\}\right], \\
        \tau &=a_{\sigma}^{q} / b_{\sigma}^{q}, \\
        \eta_{i} &=\operatorname{logit}\left(p_{0}\right)-0.5 \tau\left(\left(\mu_{i}^{q}\right)^{2}+\Sigma_{k, k}^{q}\right) \boldsymbol{h}_{i}^{T} \boldsymbol{h}_{i}+\tau \boldsymbol{h}_{i}^{T}\left[\boldsymbol{Y} \mu_{i}^{q}-\mathbf{D}_{-k} \mathbf{W}_{-i}^{q}\left(\boldsymbol{\mu}_{-i}^{q} \mu_{i}^{q}+\boldsymbol{\Sigma}_{-i, i}^{q}\right)\right], \\
        w_{i}^{q} &=\operatorname{expit}\left(\eta_{i}\right),
    \end{align}
\end{subequations}
where $\mathbf{W}^{q}=\operatorname{diag}\left(\boldsymbol{w}^{q}\right), \boldsymbol{\Omega}=\mathbf{W}^{q}\left(\mathbf{I}_{K}-\mathbf{W}^{q}\right) + \boldsymbol{w}^{q} \boldsymbol{w}^{q T}, \operatorname{logit}(A)=\ln{(A)}/\ln{(1-A)}, \operatorname{expit}(A)=\operatorname{logit}^{-1}(A)=\exp (A) (1+\exp (A))^{-1}, and \boldsymbol{w}^{q}=$ $\left[w_{1}^{q}, \ldots, w_{K}^{q}\right]^{T}$. 
To optimize the parameters of the variational distribution, an iterative coordinate-wise updating method is used. This involves initializing the variational parameters and then cyclically updating each parameter, conditioned on the updates of other parameters in the most recent iteration. 
The use of the symbol $\odot$ to represent elementwise multiplication between matrices, along with the notation $\boldsymbol{h}_{i}$ to indicate the $i^{\text {th }}$ column of matrix $\mathbf{D}$, and $\mathbf{D}_{-i}$ to denote the dictionary matrix with the $i^{\text {th }}$ column removed. 

\section{Hyperparameter setting}
To initiate the VB algorithm, the deterministic hyperparameters are initialized as follows, $v_{s}=10$, $a_{\sigma}=10^{-4}$, and $b_{\sigma}=10^{-4}$. In addition, a small probability of $p_{0}=0.1$ is assigned for the selection of a simpler model.
Initializing the variational parameters $\boldsymbol{w}^{q}$ is crucial due to the algorithm's sensitivity to their initial selection. These parameters represent the inclusion probabilities $\boldsymbol{w}^{q}$ of basis variables, which are variables used to construct a model that predicts the target variable. Inclusion probabilities represent the likelihood that each basis variable is incorporated into the model. As the algorithm progresses, it modifies these probabilities to approximate the true posterior distribution better. To initialize $\boldsymbol{w}^{q}$, FIND SINDy \cite{rudy2017data} method is used. FIND (Finding Interpretable Neural Dynamics) SINDy is a data-driven method for identifying a system's dynamics using time-series data. 

\section{Evaluation metric}
In the VB algorithm, the variational parameters are updated in every iteration, starting with the initial values until the convergence, where a convergence criterion is set as the difference between the ELBO value of two successive iterations. The updating process continues until the difference between the ELBO values of two successive iterations is less than a predefined small value $\rho$ (usually set to $10^{-6}$). At this point, the parameters are taken as optimized variational parameters, and the algorithm stops updating.
\begin{equation}
\operatorname{ELBO}^{(j)}-\operatorname{ELBO}^{(j-1)}<\rho
\end{equation}
For each iteration $j$, the value of ELBO is computed using the simplified expression.
\begin{equation}\label{eq:elbo}
    \begin{aligned}
        \mathrm{ELBO}^{(j)} &=0.5 (K- N \ln (2 \pi)- K \ln \left(v_{s}\right))-\ln \Gamma\left(a_{\sigma}\right)+\ln \Gamma\left(a_{\sigma}^{(j)}\right)+a_{\sigma} \ln \left(b_{\sigma}\right)\\
        &-a_{\sigma}^{(j)} \ln \Gamma\left(b_{\sigma}^{(j)}\right)
        +0.5 \ln \left|\boldsymbol{\Sigma}^{(j)}\right|
        +\sum_{i=1}^{K}\left[w_{i}^{(j)} \ln \left(\frac{p_{0}}{w_{i}^{(t)}}\right)+\left(1-w_{i}^{(j)}\right) \ln \left(\frac{1-p_{0}}{1-w_{i}^{(j)}}\right)\right]
    \end{aligned}
\end{equation}
The above expression come from \cite{nayek2022equation}, where the variational parameters at the $j^{\text{th}}$ iteration are denoted by $a_{\sigma}^{(j)}, b_{\sigma}^{(j)}, \boldsymbol{\mu}^{(j)}, \boldsymbol{\Sigma}^{(j)}, \boldsymbol{w}^{(j)}$. The final-stage variational parameters at convergence are denoted as, $a_{\sigma}^{*}, b_{\sigma}^{*}, \mu^{*}, \Sigma^{*}, \boldsymbol{w}^{*}$, and $\Gamma(\cdot)$ represents the Gamma function.


\begin{thebibliography}{10}

\bibitem{strauss2007partial}
Walter~A Strauss.
\newblock {\em Partial differential equations: An introduction}.
\newblock John Wiley \& Sons, 2007.

\bibitem{wazwaz2002partial}
Abdul-Majid Wazwaz.
\newblock {\em Partial differential equations}.
\newblock CRC Press, 2002.

\bibitem{helal2002soliton}
MA~Helal.
\newblock Soliton solution of some nonlinear partial differential equations and
  its applications in fluid mechanics.
\newblock {\em Chaos, Solitons \& Fractals}, 13(9):1917--1929, 2002.

\bibitem{roubivcek2013nonlinear}
Tom{\'a}{\v{s}} Roub{\'\i}{\v{c}}ek.
\newblock {\em Nonlinear partial differential equations with applications},
  volume 153.
\newblock Springer Science \& Business Media, 2013.

\bibitem{rabczuk2019nonlocal}
Timon Rabczuk, Huilong Ren, and Xiaoying Zhuang.
\newblock A nonlocal operator method for partial differential equations with
  application to electromagnetic waveguide problem.
\newblock {\em Computers, Materials \& Continua 59 (2019), Nr. 1}, pages
  31--55, 2019.

\bibitem{purohit2010fractional}
SD~Purohit and SL~Kalla.
\newblock On fractional partial differential equations related to quantum
  mechanics.
\newblock {\em Journal of Physics A: Mathematical and Theoretical},
  44(4):045202, 2010.

\bibitem{howe1980quantum}
Roger Howe.
\newblock Quantum mechanics and partial differential equations.
\newblock {\em Journal of Functional Analysis}, 38(2):188--254, 1980.

\bibitem{courant2008methods}
Richard Courant and David Hilbert.
\newblock {\em Methods of mathematical physics: partial differential
  equations}.
\newblock John Wiley \& Sons, 2008.

\bibitem{debnath2005nonlinear}
Lokenath Debnath and Lokenath Debnath.
\newblock {\em Nonlinear partial differential equations for scientists and
  engineers}.
\newblock Springer, 2005.

\bibitem{leung2013systems}
Anthony~W Leung.
\newblock {\em Systems of nonlinear partial differential equations:
  applications to biology and engineering}, volume~49.
\newblock Springer Science \& Business Media, 2013.

\bibitem{brunton2016discovering}
Steven~L Brunton, Joshua~L Proctor, and J~Nathan Kutz.
\newblock Discovering governing equations from data by sparse identification of
  nonlinear dynamical systems.
\newblock {\em Proceedings of the National Academy of Sciences},
  113(15):3932--3937, 2016.

\bibitem{mangan2016inferring}
Niall~M Mangan, Steven~L Brunton, Joshua~L Proctor, and J~Nathan Kutz.
\newblock Inferring biological networks by sparse identification of nonlinear
  dynamics.
\newblock {\em IEEE Transactions on Molecular, Biological and Multi-Scale
  Communications}, 2(1):52--63, 2016.

\bibitem{hoffmann2019reactive}
Moritz Hoffmann, Christoph Fr{\"o}hner, and Frank No{\'e}.
\newblock Reactive sindy: Discovering governing reactions from concentration
  data.
\newblock {\em The Journal of chemical physics}, 150(2):025101, 2019.

\bibitem{bhadriraju2019machine}
Bhavana Bhadriraju, Abhinav Narasingam, and Joseph Sang-Il Kwon.
\newblock Machine learning-based adaptive model identification of systems:
  Application to a chemical process.
\newblock {\em Chemical Engineering Research and Design}, 152:372--383, 2019.

\bibitem{loiseau2018constrained}
Jean-Christophe Loiseau and Steven~L Brunton.
\newblock Constrained sparse galerkin regression.
\newblock {\em Journal of Fluid Mechanics}, 838:42--67, 2018.

\bibitem{loiseau2018sparse}
Jean-Christophe Loiseau, Bernd~R Noack, and Steven~L Brunton.
\newblock Sparse reduced-order modelling: sensor-based dynamics to full-state
  estimation.
\newblock {\em Journal of Fluid Mechanics}, 844:459--490, 2018.

\bibitem{lai2019sparse}
Zhilu Lai and Satish Nagarajaiah.
\newblock Sparse structural system identification method for nonlinear dynamic
  systems with hysteresis/inelastic behavior.
\newblock {\em Mechanical Systems and Signal Processing}, 117:813--842, 2019.

\bibitem{li2019discovering}
Shanwu Li, Eurika Kaiser, Shujin Laima, Hui Li, Steven~L Brunton, and J~Nathan
  Kutz.
\newblock Discovering time-varying aerodynamics of a prototype bridge during
  vortex-induced vibrations.
\newblock In {\em APS Division of Fluid Dynamics Meeting Abstracts}, pages
  P14--007, 2019.

\bibitem{kaiser2018sparse}
Eurika Kaiser, J~Nathan Kutz, and Steven~L Brunton.
\newblock Sparse identification of nonlinear dynamics for model predictive
  control in the low-data limit.
\newblock {\em Proceedings of the Royal Society A}, 474(2219):20180335, 2018.

\bibitem{schaeffer2020extracting}
Hayden Schaeffer, Giang Tran, Rachel Ward, and Linan Zhang.
\newblock Extracting structured dynamical systems using sparse optimization
  with very few samples.
\newblock {\em Multiscale Modeling \& Simulation}, 18(4):1435--1461, 2020.

\bibitem{stender2019recovery}
Merten Stender, Sebastian Oberst, and Norbert Hoffmann.
\newblock Recovery of differential equations from impulse response time series
  data for model identification and feature extraction.
\newblock {\em Vibration}, 2(1):25--46, 2019.

\bibitem{rudy2017data}
Samuel~H Rudy, Steven~L Brunton, Joshua~L Proctor, and J~Nathan Kutz.
\newblock Data-driven discovery of partial differential equations.
\newblock {\em Science Advances}, 3(4):e1602614, 2017.

\bibitem{hao2021data}
Jiang Hao, Wang Bofu, and Lu~Zhiming.
\newblock Data-driven sparse identification of governing equations for fluid
  dynamics.
\newblock {\em Chinese Journal of Theoretical and Applied Mechanics},
  53(6):1543--1551, 2021.

\bibitem{ai2022study}
Jiali Ai, Jindong Dai, Jianmin Liu, Chi Zhai, and Wei Sun.
\newblock Study on the kinetic parameters of crystallization process modelled
  by partial differential equations.
\newblock In {\em Computer Aided Chemical Engineering}, volume~49, pages
  1099--1104. Elsevier, 2022.

\bibitem{naikdiscovering}
Richa~R Naik, Armi Tiihonen, Janak Thapa, Clio Batali, Shijing Sun, Zhe Liu,
  and Tonio Buonassisi.
\newblock Discovering the underlying equations governing perovskite solar-cell
  degradation using scientific machine learning.

\bibitem{long2019pde}
Zichao Long, Yiping Lu, and Bin Dong.
\newblock Pde-net 2.0: Learning pdes from data with a numeric-symbolic hybrid
  deep network.
\newblock {\em Journal of Computational Physics}, 399:108925, 2019.

\bibitem{fuentes2021equation}
R~Fuentes, R~Nayek, P~Gardner, N~Dervilis, T~Rogers, K~Worden, and EJ~Cross.
\newblock Equation discovery for nonlinear dynamical systems: a bayesian
  viewpoint.
\newblock {\em Mechanical Systems and Signal Processing}, 154:107528, 2021.

\bibitem{tripura2023sparse}
Tapas Tripura and Souvik Chakraborty.
\newblock A sparse bayesian framework for discovering interpretable nonlinear
  stochastic dynamical systems with gaussian white noise.
\newblock {\em Mechanical Systems and Signal Processing}, 187:109939, 2023.

\bibitem{tripura2022model}
Tapas Tripura and Souvik Chakraborty.
\newblock Model-agnostic stochastic model predictive control.
\newblock {\em arXiv preprint arXiv:2211.13012}, 2022.

\bibitem{tripura2023probabilistic}
Tapas Tripura, Aarya~Sheetal Desai, Sondipon Adhikari, and Souvik Chakraborty.
\newblock Probabilistic machine learning based predictive and interpretable
  digital twin for dynamical systems.
\newblock {\em Computers \& Structures}, 281:107008, 2023.

\bibitem{mathpati2023mantra}
Yogesh~Chandrakant Mathpati, Kalpesh~Sanjay More, Tapas Tripura, Rajdip Nayek,
  and Souvik Chakraborty.
\newblock Mantra: A framework for model agnostic reliability analysis.
\newblock {\em Reliability Engineering \& System Safety}, page 109233, 2023.

\bibitem{liu2021automated}
Ruixian Liu, Michael~J Bianco, and Peter Gerstoft.
\newblock Automated partial differential equation identification.
\newblock {\em The Journal of the Acoustical Society of America},
  150(4):2364--2374, 2021.

\bibitem{leveque2007finite}
Randall~J LeVeque.
\newblock {\em Finite difference methods for ordinary and partial differential
  equations: steady-state and time-dependent problems}.
\newblock SIAM, 2007.

\bibitem{thomas2013numerical}
James~William Thomas.
\newblock {\em Numerical partial differential equations: finite difference
  methods}, volume~22.
\newblock Springer Science \& Business Media, 2013.

\bibitem{barthelmann2000high}
Volker Barthelmann, Erich Novak, and Klaus Ritter.
\newblock High dimensional polynomial interpolation on sparse grids.
\newblock {\em Advances in Computational Mathematics}, 12:273--288, 2000.

\bibitem{bruno2012numerical}
O~Bruno and D~Hoch.
\newblock Numerical differentiation of approximated functions with limited
  order-of-accuracy deterioration.
\newblock {\em SIAM Journal on Numerical Analysis}, 50(3):1581--1603, 2012.

\bibitem{nayek2021spike}
Rajdip Nayek, Ramon Fuentes, Keith Worden, and Elizabeth~J Cross.
\newblock {On spike-and-slab priors for Bayesian equation discovery of
  nonlinear dynamical systems via sparse linear regression}.
\newblock {\em Mechanical Systems and Signal Processing}, 161:107986, 2021.

\bibitem{joyce2011kullback}
James~M Joyce.
\newblock Kullback-leibler divergence.
\newblock In {\em International encyclopedia of statistical science}, pages
  720--722. Springer, 2011.

\bibitem{nayek2022equation}
Rajdip Nayek, Keith Worden, and Elizabeth~J Cross.
\newblock Equation discovery using an efficient variational {B}ayesian approach
  with spike-and-slab priors.
\newblock In {\em Model Validation and Uncertainty Quantification, Volume 3},
  pages 149--161. Springer, 2022.

\end{thebibliography}

\end{document}